\documentclass[10pt,twocolumn,letterpaper]{article}
\usepackage[pagenumbers]{cvpr} 
\usepackage[dvipsnames]{xcolor}

\usepackage{url}
\usepackage{graphicx}
\usepackage{amsmath}
\usepackage{amssymb}
\usepackage{booktabs}
\usepackage{physics}
\usepackage{float}
\usepackage{algorithmic}
\usepackage{algorithm}

\newtheorem{problem}{Problem}

\definecolor{cvprblue}{rgb}{0.21,0.49,0.74}
\usepackage[pagebackref,breaklinks,colorlinks,citecolor=cvprblue]{hyperref}
\def\confYear{2024}
\title{Identifying Important Group of Pixels using Interactions}

\author {
    Kosuke Sumiyasu,\thanks{Graduate School of Science and Engineering, Chiba University (email: kosuke.sumiyasu@gmail.com)} 
    \ Kazuhiko Kawamoto,\thanks{Graduate School of Engineering, Chiba University (email: kawa@faculty.chiba-u.jp, kera@chiba-u.jp)}
    \ Hiroshi Kera$^{\dagger}$\thanks{Corresponding Author.} 
}

\begin{document}
\maketitle
\begin{abstract}
To better understand the behavior of image classifiers, it is useful to visualize the contribution of individual pixels to the model prediction.
In this study, we propose a method, MoXI~(\textbf{Mo}del e\textbf{X}planation by \textbf{I}nteractions), that efficiently and accurately identifies a group of pixels with high prediction confidence.
The proposed method employs game-theoretic concepts, Shapley values and interactions, taking into account the effects of individual pixels and the cooperative influence of pixels on model confidence.
Theoretical analysis and experiments demonstrate that our method better identifies the pixels that are highly contributing to the model outputs than widely-used visualization by Grad-CAM, Attention rollout, and Shapley value. While prior studies have suffered from the exponential computational cost in the computation of Shapley value and interactions, we show that this can be reduced to quadratic cost for our task.
The code is available at \url{https://github.com/KosukeSumiyasu/MoXI}.
\end{abstract}

\section{Introduction}\label{sec:introduction}

Visualization of important image pixels has been widely used to understand machine learning models in computer vision tasks such as image classification~\cite {zhou2016learning, Selvaraju2018Grad_CAM, abnar2020quantifying, Petsiuk2018rise,  Lundberg2017aunified, Binder2016layer}.
To this end, visualization methods compute the contribution of each pixel to model decisions. 
For example, Grad-CAM~\cite{Selvaraju2018Grad_CAM} measures the contribution using a weighted sum of the feature maps of convolutional layers, where weights are determined by the gradient of confidence score for any target class with respect to the feature map entries. Attention rollout~\cite{abnar2020quantifying} measures it based on the attention weight of encoders of a Vision Transformer.

\begin{figure*}[t]
\begin{center}
\includegraphics[clip, keepaspectratio, width=\textwidth]{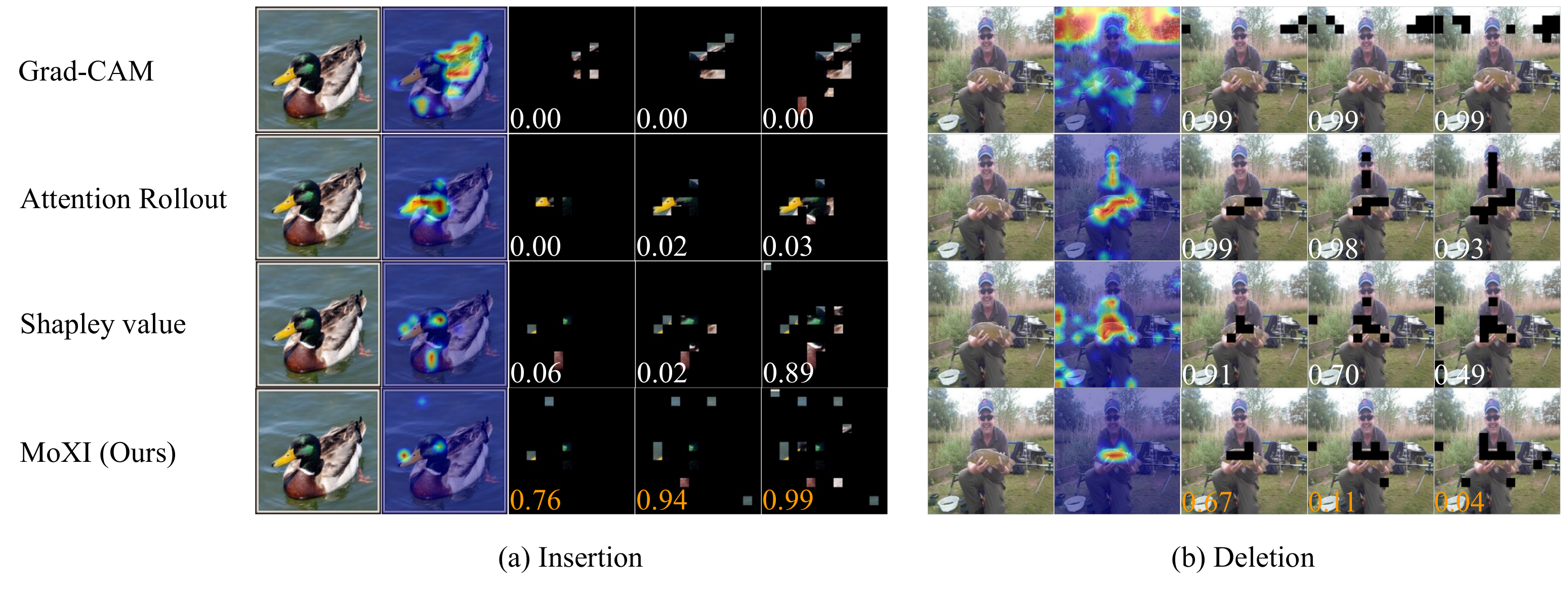}
\caption{Examples of image patches with high contributions to the output of ViT-T. (a) Starting from an empty image, image patches are inserted according to their contribution measured by each method. (b) Starting from an original image, image patches are removed according to their contribution measured by each method. 
The heatmaps highlight the image patches inserted (deleted) to obtain the correct (incorrect) classification. 
The selected patches are colored according to the timing of insertion/deletion. For insertion, only the proposed method selects patches from the background. For deletion, the proposed method highlights the class object only. 
For both cases, the proposed method highlights the least number of patches while achieving the highest/lowest confidence score. 
}
\label{fig:example}
\end{center}
\end{figure*}

Several recent studies revealed that a game-theoretic concept, \textit{Shapley values}~\cite{shapley1953contibutions}, is a powerful indicator of pixel contribution~\cite{Lundberg2017aunified, jethani2022fastshap, covert2023learning}.
In multi-player games, Shapley values measures the contribution of each player from the average change in the total game reward with his/her presence versus absence. When applied to an image classifier, the pixels of an image are the players, which work cooperatively for the model output (e.g., confidence score). Unlike Grad-CAM and Attention rollout, Shapley values compute the contribution of pixels to the model output more directly. The former methods use feature maps or attention weights, the magnitude of whose entries are not necessarily well-aligned with their contributions to the model output, whereas the latter uses logits or confidence scores. Indeed, Fig.~\ref{fig:example} shows that the pixels with high Shapely values have a significantly larger impact on confidence scores than those determined by Grad-CAM or Attention rollout in both (a) insertion case and (b) deletion case. 

A crucial caveat of the aforementioned methods is that they identify a group of important pixels by the \textit{individual} contribution of each pixel and overlook the collective contribution of multiple pixels. For example, Fig.~\ref{fig:example}(a) shows that the three methods only highlight the class object (i.e., duck) and do not indicate the background (i.e., sea) as an informative factor. 
However, the set of pixels with the highest contributions (e.g., highest Shapley values) does not imply the most informative pixel set as a whole because the information overlap among pixels is not considered. Indeed, the bottom row of Fig.~\ref{fig:example}(a) shows that the class object and background greatly impact in synergy the confidence score.

In this paper, we propose an efficient game-theoretic visualization method of image pixels with a high impact on the prediction of an image classifier. Besides Shapley values, we exploit \textit{interactions}, a game-theoretical concept that reflects the average effect of the cooperation of pixels. Namely, unlike prior methods, including Grad-CAM, Attention rollout, and Shapley values, the proposed method takes into account the cooperative contribution of pixels and identifies the image pixels as a whole. In Fig.~\ref{fig:example}(a), the proposed method identifies a pixel set on which the classifier puts high classification confidence. Similarly, in Fig.~\ref{fig:example}(b), it identifies a minimal pixel set without which the classification fails.
Notably, we define \textit{self-context} variants of Shapley values and interactions, and reduce the number of forward passes from exponential to quadratic times, which resolves the fundamental challenge of game-theoretic approaches to be handy tools for model explanation. 

In the experiments, we consider the insertion curve and deletion curve on a subset of ImageNet images that are correctly classified by a pretrained classifier. Starting from fully masked images, an insertion curve plots the increase of classification accuracy as unmasking image patches from highly contributing ones determined by each method. Similarly, a deletion curve plots the accuracy decrease from the clean images to fully masked ones. The results show that the proposed method gives sharp insertion/deletion curves. For example, the classification accuracy reached $90\%$ with images with 4\% unmasked patches if selected by the proposed method, significantly outperforming the results of Grad-CAM~(accuracy of $2\%$), Attention rollout~(accuracy of $4\%$), and Shapley values~(accuracy of $25\%$).
Similar results are observed for the deletion curves and also when we use common corruptions~\cite{hendrycks2018benchmarking} instead of masking.
Qualitatively, the heatmaps using the patches selected in the early stage of the insertion curve show that our method highlights both a class object and background, while the other methods mostly highlight the class object only. Meanwhile, in the heatmaps from the deletion curves, our method particularly highlights the class-discriminative region of the object, while the others do not. 

Our contributions are summarized as follows:
\begin{itemize}
    \item We propose an efficient game-theoretic visualization method, named MoXI~(\textbf{Mo}del e\textbf{X}planation by \textbf{I}nteractions), for a group of pixels that significantly influences the classification.
    \item Our analysis supports a simple greedy strategy from a game-theoretic perspective, leading us to use self-context variants of Shapley values and interactions, which can be computed exponentially faster than computing the original ones.
    \item Extensive experiments show that our method more accurately identifies the pixels that are highly contributing to the model outputs than standard visualization methods.
\end{itemize}

\section{Related Work}\label{sec:related_work}
\paragraph{Visual explanation of model decision.}
Various methods have been proposed to understand deep learning models for vision tasks by quantifying and visualizing the contribution of image pixels to the model output~\citep{zhou2016learning, Selvaraju2018Grad_CAM, abnar2020quantifying, Petsiuk2018rise,  Lundberg2017aunified, Binder2016layer,chattopadhyay2018grad_cam,wang2020scorecam, Chefer2021TransformerInt}.
The contribution of pixels has been typically measured using feature maps in models.
For example, Grad-CAM~\citep{Selvaraju2018Grad_CAM} determines the contribution by applying weights to the feature maps of the convolutional layers of a CNN using gradients.
Attention rollout~\citep{abnar2020quantifying}, commonly used for Vision Transformers, calculates the contributions using attention maps. 
Several methods instead calculate the contribution of each pixel by analyzing the sensitivity of the confidence score with respect to each pixel~\citep{Petsiuk2018rise, Lundberg2017aunified, jethani2022fastshap, covert2023learning}. For example, RISE~(Randomized Input Sampling for Explanation;~\citep{Petsiuk2018rise}) calculates the contributions empirically by probing the model with randomly masked images of the input image and obtaining the corresponding confidence scores. 
SHAP~(SHapley Additive exPlanations;~\citep{Lundberg2017aunified}) distributes confidence scores fairly to contributions by leveraging Shapley values from game theory.
Importantly, the aforementioned methods all measure the contribution of each pixel independently; the collection of important pixels consists of the pixels with high contributions. In contrast, this study identifies the important pixels by further taking into account the collective contributions of pixels.

\paragraph{Game-theoretic approach of model explanation.}
Several recent studies have utilized a game-theoretic concept, interactions, to analyze various phenomena of deep learning models and quantify an effect of pixel cooperation on the model inference~\citep{cheng2021agame, deng2022discovering, ren2021a, wang2021a, zhang2021interpreting,sumiyasu2022gametheoretic}.\footnote{There is also a similar approach to measure the second-order effects~\cite{blucher2022prediff,bluecher2024decoupling}, but this paper focuses on Shapley values and interactions, which are more widely used.}
\citet{wang2021a} showed that the transferability of adversarial images has a negative correlation to the interactions. 
\citet{zhang2021interpreting} showed the similarity between the computation of interactions and dropout regularization.
\citet{deng2022discovering} discussed the difference in information obtained between humans and machine learning models using interactions.
\citet{sumiyasu2022gametheoretic} investigated misclassification by models using interactions and discovered that the distribution of interactions varies with the type of misclassified images.
Thus, interactions are helpful for understanding the model from the perspective of cooperative relationships between pixels.
A critical issue of interaction-based analysis is its computational cost; the computation of interaction requires an exponential number of forward passes with respect to the number of pixels.
In this paper, we propose an efficient approach to explain a model using variants of interactions (and also Shapley values), achieving the identification of important pixels with only a quadratic number of forward passes.

\section{Preliminaries}\label{sec:preliminary}

\paragraph{Shapley values.} Shapley values was proposed in game theory to measure the contribution of each player to the total reward that is obtained from multiple players working cooperatively~\citep{shapley1953contibutions}. 
Let $N = \{1, \ldots, n\}$ be the index set of players, and let $2^N \stackrel{\rm{def}}{=} \{S\,|\, S\subseteq N\}$ be its power set. 
Given a reward function $f: 2^N \to \mathbb{R}$, the Shapley value $\phi(i\,|\, N)$ of player $i$ with a context $N$ is defined as follows.
\begin{align}\label{equ:shapley}
    \phi(i\,|\, N)\stackrel{\rm{def}}{=} \sum_{S\subseteq N \setminus \{i\}}P(S\,|\, N \setminus\{i\} ) \:[f(S\cup\{i\})-f(S)],
\end{align}
where $P~(A\,|\, B) = \frac{(\abs{B} -  \abs{A})!\abs{A} !}{(\abs{B} + 1)!}$. 
Here, $\abs{\,\cdot\,}$ denotes the cardinality of set.
Namely, the Shapley value $\phi(i\,|\, N)$ averages over all $S\subseteq N \setminus \{i\}$ the reward increase on the participation of player $i$ to player set $S$.

\paragraph{Interactions.} Interactions measure the contribution of the cooperation between the two players to the total reward~\citep{grabisch1999an}. 
Interactions $I(i,j)$ by players $i$ and $j$ are defined as follows.
\begin{align}\label{equ:interaction}
    I(i,j \,|\, N)\stackrel{\rm{def}} {=}& \phi(S_{ij} \,|\, N')-\phi(i \,|\, N \setminus \{j\} )-\phi(j \,|\, N \setminus \{i\} ),
\end{align}
where two players $i,j \in N$ are regarded as a single player $S_{ij}=\{i,j\}$ and $N' = N \setminus \{i,j\} \cup \{S_{ij}\}$~(i.e., $\abs{N'} = n-1$). 
In Eq.~\eqref{equ:interaction}, the first term corresponds to the joint contribution from players $(i, j)$, and the second and the third terms correspond to the individual contribution of players $i$ and $j$, respectively. Namely, interactions quantify the average cooperation on the reward of two players joining \textit{simultaneously}. Importantly, we have $I(i,i \,|\, N) = -\phi(i\,|\, N)$.

\paragraph{Application to image classifiers.} In the application of Shapley values and interactions to image classifiers, 
an image $x$ with $n$ pixels is regarded as the index set $N = \{1, \ldots, m\}$ of players. 
Typically, the reward function $f$ is defined by $f(x) = \log \frac{P(y \,|\, x)}{1-P(y \,|\, x)}$~\citep{deng2022discovering}, where $y$ represents the class of $x$, and $P(y \,|\, x)$ denotes the classifier's confidence score on class $y$ with input $x$. 
The reward $f(S)$ of a subset of pixels $S\subset 2^N$ of image $x$ is similarly computed by feeding a partially masked $x$ to the classifier~(i.e., the pixels in $N\setminus S$ are masked).

If the classifier is a convolutional neural network~(CNN), the masked region is conventionally filled with some base value, such as 0 or the average pixel value~\cite{ancona19a, zhang2020Interpreting}. Such a replacement may drop the original information of an image but also inject a new feature. Thus, the choice of base value affects the Shapley values and interactions. In contrast, when a Vision Transformer is used, one can realize masking in a rigid manner by applying a mask to the attention. To our knowledge, most prior studies exploited Shapley values and interactions on CNNs with the base value replacement, which might not unleash the full potential of these quantities. To our knowledge, the only exception is~\cite{covert2023learning}, which demonstrated that Shapley values can be calculated more accurately using attention masking. We follow this strategy in the computation of Shapley values and interactions for Vision Transformers.

\section{Method}\label{sec:method}

We address the problem of identifying in a given image a set of pixels that significantly influence the confidence score of a classifier. While prior studies solve this by explicitly or implicitly measuring the independent contribution of each pixel to the confidence score, the proposed method takes into account the collective contribution of pixels using interactions. We refer to the proposed method as MoXI~(\textbf{Mo}del e\textbf{X}planation by \textbf{I}nteractions). 

We consider two approaches to measuring the contribution of pixels to the confidence score: (i) pixel insertion and (ii) pixel deletion. The former measures the contribution of a pixel by the confidence gain when it is unmasked as in Eqs.~\eqref{equ:shapley} and~\eqref{equ:interaction}, while the latter measures it by the confidence drop when it is masked.

\subsection{Pixel Insertion}\label{subsec:method_insert}
\begin{problem}\label{problem_1}
Let $N$ be the index set of all pixels of image $x$. 
Let $f:2^N \to [0, 1]$ be a function that gives the confidence score on the class of index set, with the convention that pixels not included in the index set are masked. 
Find a subset $S_k \subset N$ such that 
\begin{align}
    S_k = \underset{{S\subseteq N, |S|=k}}{\mathrm{arg\,max}}\ \ f(S),
\end{align}
for $k = 1, 2, \ldots, |N|$.
\end{problem}
By its formulation, this problem is an NP-hard problem in general. Particularly, $f$ is here a CNN or Vision Transformer,\footnote{With this assumption, we use a slight abuse of notation and assume, e.g., $f(\{a, \{b, c\}\}) = f(\{a, b, c\})$ because in either case of $\{a, \{b, c\}\}$ or $\{a, b, c\}$, we input the image with pixels $a, b, c$ to the model. } a highly nonlinear function. Thus, we resort to a greedy strategy to solve it approximately.  

For $k=1$, the index $b_1 \in N$ of the pixel with the highest Shapley value of $\phi(b_1\,|\, \{b_1\})$ gives the optimal set $S_1 = \{b_1\}$ by the its definition. 
For $k=2$, we select the next pixel $b_2$ with the one maximizing $f(\{b_1, b_2\})$. Importantly, this is equivalent to maximizing the sum of the Shapley value and interaction, not the Shapley value alone. 
\begin{align}\label{equ:optimial-S2}
 b_2 
 &= \underset{b\in N\setminus \{b_1\}}{\mathrm{arg\,max}} \ f(\{b_1, b\}) - f(\varnothing)\nonumber\\ 
 &= \underset{b\in N\setminus \{b_1\}}{\mathrm{arg\,max}} \ \phi(\{b_1, b\} \,|\, \{\{b_1, b\}\})\nonumber \\ 
 &= \underset{b\in N\setminus \{b_1\}}{\mathrm{arg\,max}}\ \phi(b \,|\, \{b\}) + I(b_1, b \,|\, \{b_1, b\}) \nonumber\\
 &= \underset{b\in N\setminus \{b_1\}}{\mathrm{arg\,max}}\ \phi^{(0)}(b) + I^{(0)}(b_1, b),
\end{align}
where 
\begin{align}
  \phi^{(0)}(a) &\stackrel{\rm{def}}{=} \phi(a \,|\, \{a\}) = f(a) - f(\varnothing) \\ 
  I^{(0)}(a_1, a_2) &\stackrel{\rm{def}}{=} I(a_1, a_2 \,|\, \{a_1, a_2\}) \nonumber\\
  &= f(a_1\cup a_2) - f(a_1) - f(a_2) + f(\varnothing).
\end{align}
We refer to such a particular form of Shapley values and interactions to be \textit{self-context} in the pixel insertion approach, and they play an essential role in our framework.  
For $k \ge 3$, we can similarly show that maximizing $f(S_{k-1}\cup \{b_k\})$ with respect to $b_k$ is equivalent to 
\begin{align}\label{equ:optimal-Sk}
    b_k 
    &= \underset{b \in N \setminus S_{k-1}} {\operatorname{argmax}} \:\phi^{(0)}(b) + I^{(0)}(S_{k-1}, b).
\end{align}
Equation~\eqref{equ:optimal-Sk} shows that for identifying of index $b_k$ for $S_k$, it is crucial to consider the interaction between $S_{k-1}$ and $b_k$. 
Even when a pixel indexed $b$ has a large Shapley value~(the first term), it may have a large negative interaction~(the second term) if its pixel information overlaps with that of $S_{k-1}$. Namely, collecting pixels with large Shapley values does not necessarily give the most informative pixel set. 

To summarize, our analysis justifies a very simple greedy algorithm~Algorithm~\ref{alg:insert} from a game-theoretic perspective. The algorithm seems trivial in hindsight, but prior studies visualize highly contributing pixels only using Shapley values~\cite{Lundberg2017aunified, jethani2022fastshap, covert2023learning}. 

\begin{figure}[t]
\begin{algorithm}[H]
    \caption{Identification of a group of pixels in the pixel insertion approach}
    \label{alg:insert}
    \begin{algorithmic}[1]
    \REQUIRE reward function $f$, index set $N$ of image pixels.
    \ENSURE Sequence of subsets $S_1, \ldots, S_{|N|}\subset N$
    \STATE $S_k \leftarrow \{\} \; \text{for all} \; k=0,\ldots,|N|$
    \FOR{$k=1, \ldots, |N|$}
        \STATE $b_k \leftarrow \underset{b \in N \setminus S_{k-1}} {\operatorname{argmax}} f(S_{k-1}\cup \{b\})$
        \STATE $S_k \leftarrow S_{k-1} \cup \{b_k\}$
    \ENDFOR
    \RETURN $S_1, \ldots, S_{|N|}$
    \end{algorithmic}
\end{algorithm}
\end{figure}

\paragraph{Computational cost.}
The identification of important pixels (or patches, in practice) using Shapley values requires $\mathcal{O}(|N|2^{|N|})$ times of forward passes because of the average over all $S \in N\setminus \{i\}$ for all $i\in N$~(cf.~Eq.~\eqref{equ:shapley}). 
In contrast, our approach only requires $\mathcal{O}(|N|^2)$ times of forward passes in the worst case (see Appendix~\ref{app:runtime} for details of the algorithm complexity and runtime).

\paragraph{\textbf{\textsc{Set-Sum}} task.}
We now give an intuitive example for showing the necessity of interactions using \textsc{Set-Sum} task. 
\textsc{Set-Sum} task is a variant of Problem~\ref{problem_1} with a collection of integers $N \subset \mathbb{Z}$ and reward function $f(S) =s$ for $S \subseteq N$, where $s$ denotes the sum of \textit{all types of} integers in $S$. 
For example, $s = 3$ for $S=\{2, 2, 1\}$.
Note that for any $i \in N$, we have $f(S_{k-1}\cup \{i\}) = f(S_{k-1}) + i$ if $i \notin S_{k-1}$ and otherwise $f(S_{k-1}\cup \{i\}) = f(S_{k-1})$.
In this way, when the features already possessed are equal to the newly added features, the model does not gain new information.
This shows the role of interaction in considering information redundancy.

\begin{figure}[t]
    \begin{center}
    \includegraphics[clip, keepaspectratio, width=\columnwidth]{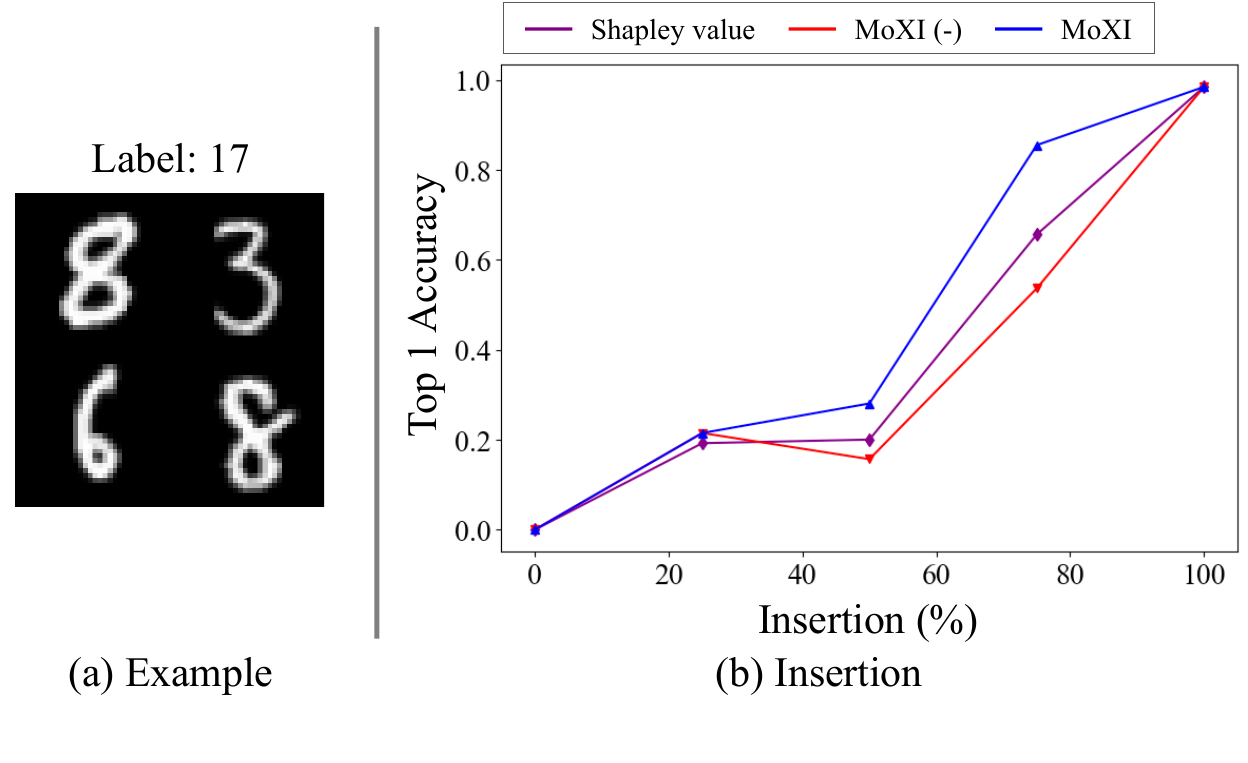}
    \caption{(a) Example of a synthetic MNIST image in the visual \textsc{Set-Sum} task, labeled 17 by the sum of all types of numbers in the image.
    (b) Insertion curves. The curves illustrate the change of accuracy when adding image patches gradually with high contributions identified by different methods at various unmasked image rates, ranging from $0$ to $100\%$.
    These curves use a masking method that fills in zeros for game-theoretic calculations and model input during classification accuracy measurement.
    MoXI(-) only employs self-context Shapley values, whereas MoXI additionally uses interactions across highly contributing patches.
    }
    \label{fig:mnist_inclusion}
    \end{center}
\end{figure}

\paragraph{Visual \textbf{\textsc{Set-Sum}} task.} We empirically confirm the advantage of using interactions in the visual \textsc{Set-Sum} task on the synthetic MNIST dataset.
This task is to accurately predict the sum of all types of numbers in an image using a model.
We constructed composite images, each of which consists of four randomly selected MNIST images arranged in a 2x2 grid~(cf. Fig.~\ref{fig:mnist_inclusion}(a)). 
The label of a composite image is the sum of all types of numbers in the image as in the \textsc{Set-Sum} problem. 
The evaluation metric utilizes the insertion curve, as detailed in Sec.~\ref{sec:experiments}.
For the model and dataset details, refer to Appendix~\ref{app:visual_setsum}.
The insertion curves in Fig.~\ref{fig:mnist_inclusion}(b) show that the MoXI achieves higher accuracy than the methods using MoXI(-), which uses self-context Shapley values, and the Shapley value methods when $50\%$ and $75\%$ of the image area are unmasked, i.e., the second and the third number is appended. 
This demonstrates that MoXI acquires non-redundant information more effectively.

\subsection{Pixel Deletion}

To address Problem~\ref{problem_1}, we considered the problem of identifying groups of pixels with high confidence scores through pixel insertion. Here, we aim at decreasing the confidence scores via pixel deletion.
\begin{problem}\label{problem_2}
With the same conditions as outlined in Problem~\ref{problem_1},
find a subset $S_k \subset N$ such that 
\begin{align}
    S_k = \underset{{S\subseteq N, |S|=k}}{\mathrm{arg\,min}}\ \ f(N\setminus S),
\end{align}
for $k = 1, 2, \ldots, |N|$.
\end{problem}
We again resort to a greedy approach. The key difference is that now we define and utilize a variant of Shapley value that measures the contribution of a player by its absence.
\begin{align}\label{equ:shapley_delete}
    \phi_{\mathrm{d}}(i\,|\, N) &\stackrel{\rm{def}}{=} \sum_{S\subseteq N, i\in S} P_{\mathrm{d}} (S \setminus \{i\} \,|\, N) \qty[f(S)-f(S\setminus \{i\})],
\end{align}
where $P_{\mathrm{d}}(A\,|\,B) = \frac{(|B| - |A| - 1)! |A|!}{|B|!}$. 
This Shapley value quantifies the average impact attributable to the removal of player $i$.
In Problem~\ref{alg:insert}, we addressed the issue by defining self-context Shapley values and interactions, as it involves the case of incrementally adding pixels from the entire image. 
In contrast, Problem~\ref{alg:delete} involves the sequential deletion of pixels from an image, necessitating the formulation of \textit{full-context} Shapley values and interactions as follows:
\begin{align}
  &\phi_{\mathrm{d}}^{(|N|)}(a) 
  \stackrel{\rm{def}}{=}P_{\mathrm{d}}(S\setminus \{a\} \,|\, N) [f(N)-f(N\setminus \{a\})]\nonumber\\
  & \quad \quad \quad \: = \frac{1}{|N|}\: [f(N)-f(N\setminus \{a\})]\\
  &I_{\mathrm{d}}^{(|N|-1)}(a_1, a_2) \nonumber\\
  &\stackrel{\rm{def}}{=} \phi_{\mathrm{d}}^{(|N|-1)}(\{a_1, a_2\} \,|\, N\setminus \{a_1, a_2\}\cup\{\{a_1, a_2\}\} ) \nonumber\\
  &\quad\quad - \phi_{\mathrm{d}}^{(|N|-1)}(a_1 \,|\, N\setminus \{a_2\}) - \phi_{\mathrm{d}}^{(|N|-1)}(a_2 \,|\, N\setminus \{a_1\})\nonumber\\
  &= \frac{1}{|N|-1} [f(N) - f(N\setminus \{a_1\}) \nonumber\\
  &\quad\quad - f(N\setminus \{a_2\}) + f(N\setminus \{a_1, a_2\})].
\end{align}
With these quantities, the greedy algorithm for pixel deletion is as follows. 
For $k=1$, the index $b_1 \in N$ of the pixel with the highest (deletion-based) Shapley value $\phi_{\mathrm{d}}^{(|N|)}(b_1) =-\frac{1}{|N|}\: [f(N\setminus \{a\}) - f(N)] $ gives the optimal set $S_1=\{b_1\}$ by its definition. 
For $k=2$, we select the next pixel $b_2$ that minimizes $f(N\setminus \{b_1, b_2\})$. This choice is again explained as a sum of Shapley value and interaction,
\begin{align}\label{equ:optimial-delete-S2}
 b_{2} 
 &= \underset{b\in N\setminus \{b_1\}}{\mathrm{arg\,min}} \ f(N \setminus \{b_{1}, b\}) - f(N)\nonumber\\ 
 &= \underset{b\in N\setminus \{b_1\}}{\mathrm{arg\,max}} \ \phi_{\mathrm{d}}^{(|N|-1)}(\{b_1, b\} \,|\, N\setminus \{b_1, b\}\cup\{\{b_1, b\}\} ) \nonumber\\ 
 &= \underset{b\in N\setminus \{b_1\}}{\mathrm{arg\,max}} \ \phi_{\mathrm{d}}^{(|N|-1)}(b \,|\, N\setminus \{b_1\}) + \phi_{\mathrm{d}}^{(|N|-1)}(b_1 \,|\, N\setminus \{b\}) \nonumber\\ 
 &\quad\quad\quad\quad\quad\quad + I_{\mathrm{d}}^{(|N|-1)}(b_{1}, b) 
\end{align}
For $k\geq3$, we can similarly show that minimizing $f(N \setminus S_{k-1}\cup \{b\})$ with respect to $b$ is equivalent to
\begin{align}\label{equ:optimal-delete-Sk}
    b_k 
    &= \underset{b \in N \setminus S_{k-1}} {\operatorname{argmax}} \:[\phi_{\mathrm{d}}^{(|N|-1)}(b \,|\, N\setminus \{S_{k-1}\}) \nonumber\\ 
    &\quad\quad\quad\quad\quad\quad+ \phi_{\mathrm{d}}^{(|N|-1)}(S_{k-1} \,|\, N\setminus \{b\}) \nonumber\\
    &\quad\quad\quad\quad\quad\quad+ I_{\mathrm{d}}^{(|N|-|S_{k-1}|)}(S_{k-1}, b)].
\end{align}
\begin{figure}[t]
\begin{algorithm}[H]
    \caption{Identification of a group of pixels in the pixel deletion approach}
    \label{alg:delete}
    \begin{algorithmic}[1]
    \REQUIRE reward function $f$, index set of all images $N$.
    \ENSURE Sequence of subsets $S_1, \ldots, S_{|N|}\subset N$
    \STATE $S_k \leftarrow \{\} \; \text{for all} \; k=0,\ldots, |N|$
    \FOR{$k=1,\ldots, |N|$}
        \STATE $b_k \leftarrow \underset{b\in N\setminus S_{k-1}} {\operatorname{argmin}} f(N \setminus (S_{k-1} \cup \{b\}))$
        \STATE $S_k \leftarrow S_{k-1} \cup \{b_k\}$
    \ENDFOR
    \RETURN $S_1, \ldots, S_{|N|}$
    \end{algorithmic}
\end{algorithm}
\end{figure}
Again, the greedy algorithm is described from a game-theoretic viewpoint. In this case, we have two Shapley value terms. Algorithm~\ref{alg:delete} summarises the procedure. 
The computational cost of the pixel deletion approach is the same as the pixel insertion approach, which only requires $\mathcal{O}(|N|^2)$ times of forward passes in the worst case.

\section{Experiments}\label{sec:experiments}
In this section, we evaluate the characteristics of identified patches through comparative experiments with existing methods
and demonstrate the effectiveness of our method.

\paragraph{Setup.}
Our experiments utilize the ImageNet dataset~\citep{deng2009imagenet} and focus on analyzing Vision Transformer~\citep{dosovitskiy2021an} pre-trained for the classification task.
For baseline methods, we use Grad-CAM~\citep{Selvaraju2018Grad_CAM}\footnote{The target layer of Grad-CAM is set to the one before the layer normalization in the final attention block of network. This choice is common, see~\url{https://github.com/jacobgil/pytorch-grad-cam}.}, Grad-CAM++~\cite{chattopadhyay2018grad_cam}, 
Attention rollout~\citep{abnar2020quantifying}, Shapley values, and MoXI(-), which do not utilize the interactions present in MoXI.
For insertion curve experiments, we use the Pixel Insertion approach, while for deletion curves, we utilize the Pixel Deletion approach.
Following the previous studies~\citep{ren2021a, zhang2021interpreting}, we consider image patches instead of pixels to reduce computational costs. 
All methods calculate the contributions for $14\times14$ patches with a patch size of $16 \times 16$, which is equal to the patch size and the number of tokens in standard ViT models.
We used a pre-trained ViT-T\footnote{\url{https://huggingface.co/WinKawaks/vit-tiny-patch16-224}}~\cite{dosovitskiy2021an}, DeiT-T\footnote{\url{https://huggingface.co/facebook/deit-tiny-patch16-224}}~\cite{touvron2021training} and ResNet-18\footnote{\url{https://huggingface.co/microsoft/resnet-18}}~\cite{he2016deep}. 
We selected $1000$ images, one corresponding to each label, all of which were successfully classified in the test set.
To reduce the computational burden, we computed Shapley values approximately by random sampling of $S$ in Eq.~\eqref{equ:shapley} as in other studies~\cite{CASTRO2009polynomial, ren2022towards, zhang2021interpreting, sumiyasu2022gametheoretic}. 
The sampling size is set to 200.
Moreover, we have adopted feature patch deletion as the masking method for Shapley values and interactions.
In the following, we focus on ViT-T. See Appendix~\ref{app:deit} for more results.

\subsection{Evaluating the importance of identified patches}\label{sec:evaluate_curve}
We evaluate the importance of the image patches as determined by the above methods, using insertion/deletion curve metrics.
The insertion curve identifies information-rich patches, while the deletion curve 
helps identify patches important for the model's decision-making process.
In our insertion/deletion curve experiments,
we utilized the masking method for patch deletion. For Grad-CAM, Attention rollout, and Shapley value, image patches are inserted and deleted 
in the same order.

\begin{figure}[t]
    \begin{center}
    \includegraphics[clip, keepaspectratio, width=\linewidth]{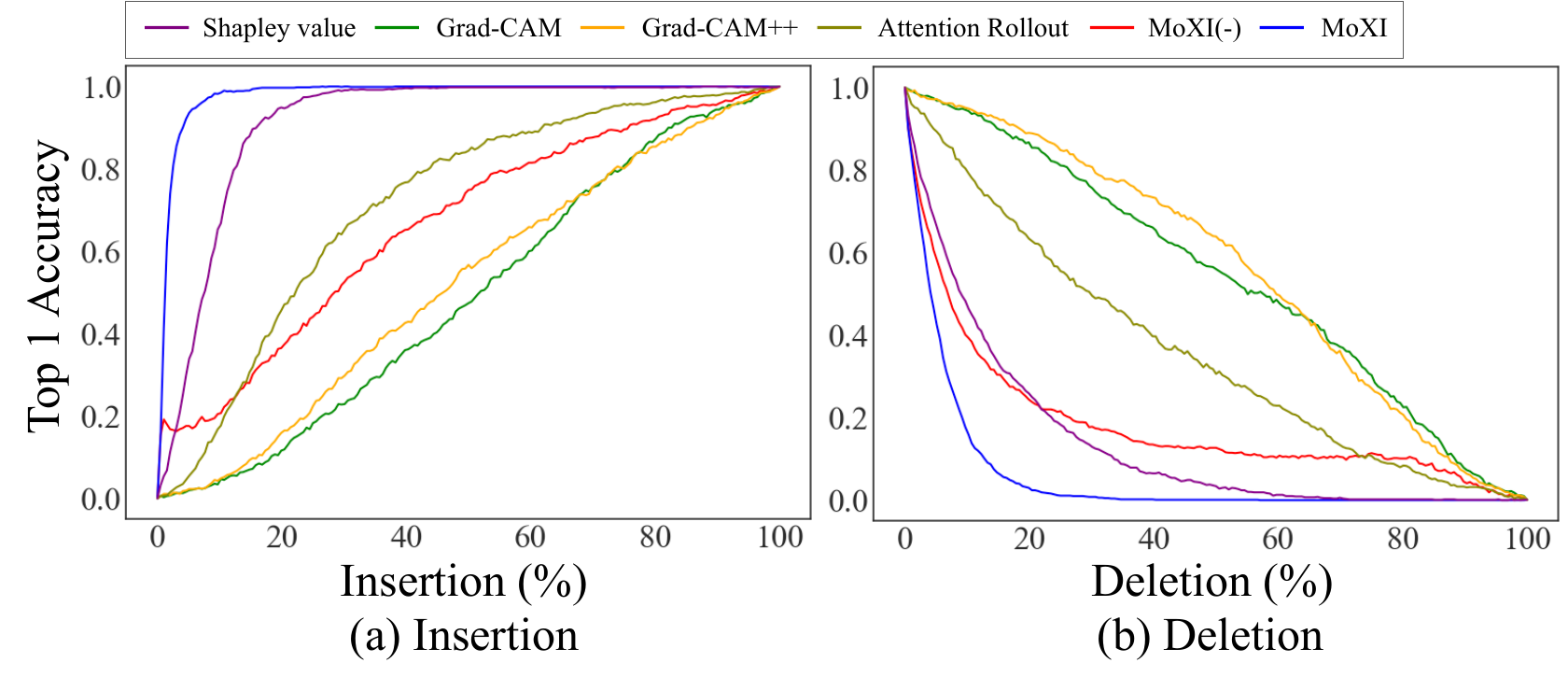}
    \caption{(a) Insertion curves. (b) Deletion curves. 
    The curves illustrate the change of accuracy when appending (removing) image patches gradually with high contributions identified by different methods at various unmasked (masked) image rates, ranging from $0$ to $100\%$.
    }
    \label{fig:imagenet_inclusion_deletion}
    \end{center}
\end{figure}

\begin{figure*}[t]
    \begin{center}
    \includegraphics[clip, keepaspectratio, width=\linewidth]{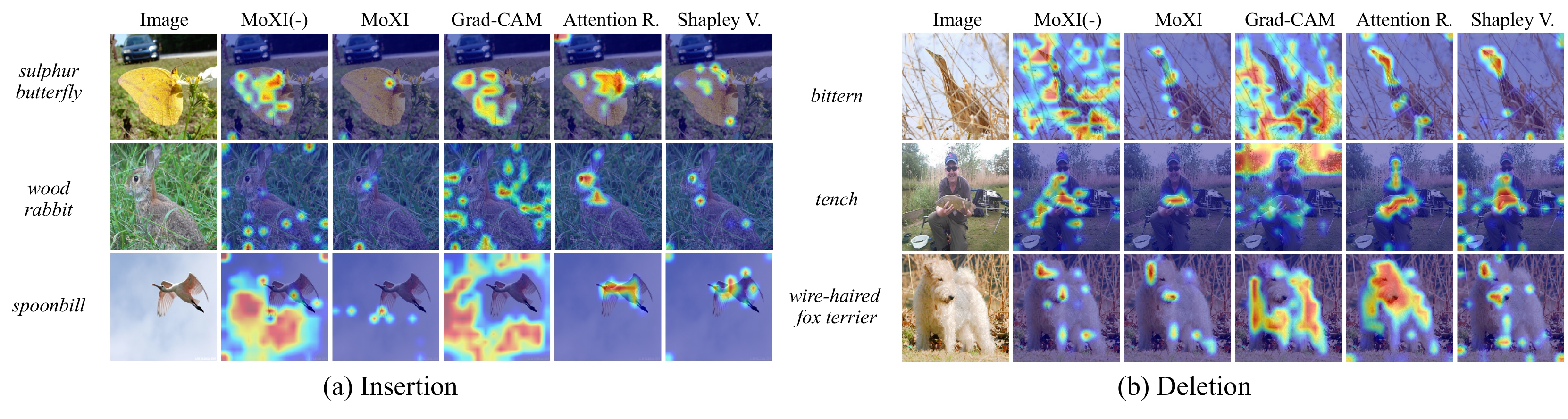}
    \caption{Visualization of important image patches by each method. 
    The highlighted image patches are selected based on their contributions calculated by each method.
    (a) Highlighting the patches incrementally added to an entire image until classification success.
    (b) Highlighting the patches sequentially removed from a full image until classification failure.
    }
    \label{fig:heatmap1}
    \end{center}
\end{figure*}

The insertion curves in Fig.~\ref{fig:imagenet_inclusion_deletion}(a) show that MoXI exhibits a sharper increase in classification accuracy compared to the other methods.
In particular, even with images where only $4\%$ is visible, MoXI achieves an accuracy of $90\%$, whereas Grad-CAM, Attention rollout, and Shapley value achieve $2\%$, $4\%$, and  $25\%$, respectively.
This result indicates that MoXI can efficiently identify important patches for classification.
Then, both the self-context and original Shapley values, which are based on confidence scores, achieve a sharper increase in classification accuracy.
However, these two methods calculate the importance of individual patches and often
select patches with similar information. 
Consequently, MoXI can identify features contributing to a higher classification accuracy than these methods.

The deletion curves in Fig.~\ref{fig:imagenet_inclusion_deletion}(b) show that MoXI exhibits a sharp decrease in classification accuracy compared to the other methods.
When concealing just $10\%$ of an image, MoXI significantly decreases the model's accuracy to $16\%$. In contrast, Grad-CAM and Attention rollout only decrease the accuracy to approximately $79\%$ under the same conditions.
This result indicates that MoXI, which accounts for interactions between patches, effectively identifies the image patches
important for classification.
We observed analogous results for DeiT-T~\cite{touvron2021training} and ResNet-18~\cite{he2016deep} models,
as detailed in Appendix~\ref{app:deit}.
Additionally, we discuss the application of masks using our method in Appendix~\ref{app:mask_layer}.

\subsection{Confidence score-based visualization}\label{subsec:visualization_method}
We introduce two heatmap-based visualization methods tailored for analyzing insertion and deletion patches.
The first method visualizes insertion patches, highlighting those important for accurate classification. 
The second focuses on deletion patches, specifically identifying those whose deletion significantly impacts the classification.
The heatmap shows higher values, indicated by shades closer to red, for patches that were inserted or deleted earlier.
The insertion or deletion stops
when the model reaches a successful classification or misclassification. 

\paragraph{Heatmap visualization.}
Figure~\ref{fig:heatmap1}(a) displays a heatmap for patch insertion.
Compared to the existing methods,
MoXI's heatmap highlights fewer regions and identifies
the class object.
Interestingly, MoXI selects the patches on the background
as well as the class object.
This visualization explains the object and background
is required for classification and 
demonstrates the usefulness of the interaction.

Figure~\ref{fig:heatmap1}(b) displays a heatmap for patch deletion. 
The heatmaps generated by MoXI(-) and Grad-CAM display extensive highlights across the image,
while MoXI, Attention rollout, and Shapley value
show more concentrated highlights on the class object. 
This finding indicates that 
these latter methods accurately
capture important information from the object.
Notably, MoXI places less emphasis on the background than Attention rollout and Shapley value. 
This result suggests that MoXI effectively narrows down information by selectively deleting the class object,
which could be advantageous for precise object localization.

\paragraph{Class-dicriminative localization.}

To enhance understanding of the model's prediction process, localization for
specific classes improve interpretability.
We have extended MoXI to analyze a target class that
differs from the model's prediction.
For the detailed visualization, see Appendix~\ref{app:visualize_specific}. Figures~\ref{fig:dog_and_cat}(b) and ~\ref{fig:dog_and_cat}(c) visualizes 
important regions for two classes: the bull mastiff, as predicted by the model, and the tiger cat, the target class. 
The heatmaps reveal that MoXI highlights the bull mastiff's facial area and the tiger cat's face and body. 
These observations demonstrate that MoXI can identify important groups of image patches relevant to the predicted class and class-specific features important for decision-making.

\begin{figure}[t]
    \begin{center}
    \includegraphics[clip, keepaspectratio, width=\linewidth]{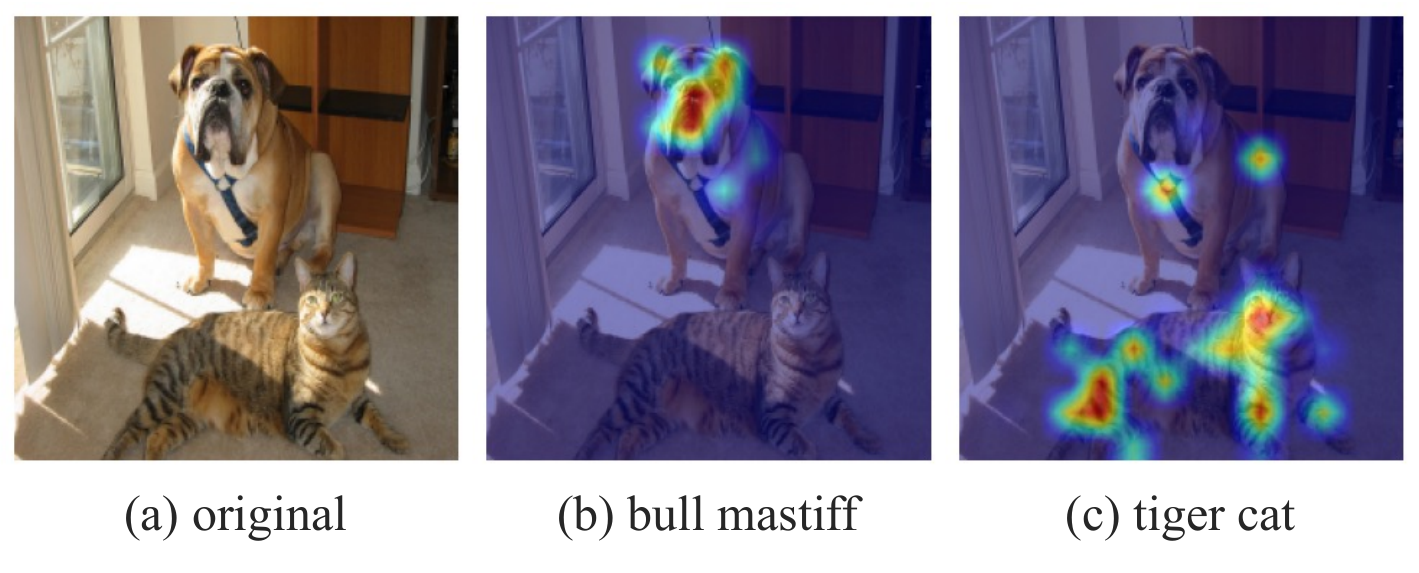}
    \caption{Visualization of important region for a targeted class using the proposed method. 
    (a) Original image. 
    (b) Targeting the bull mastiff class, which is predicted by the model. The highlighted patches are those sequentially removed from a full image until predict the bull mastiff class. 
    (c) Targeting tiger cat class. 
    We first removed the patches that has a positive contribution to bull mastiff class and also negative contribution to tiger cat. Once the tiger cat becomes the predicted class of the model, the patches highly contributing to tiger cat is removed sequentially until the prediction change, which are the highlighted patches. 
    }
    \label{fig:dog_and_cat}
    \end{center}
\end{figure}

\subsection{Common corruption effect on patch deletion}\label{subsec:realistic}

We investigate the risk of model misclassification when 
image patches important for model accuracy are disrupted
by adding noise.
In the deletion curve experiment of Sec.~\ref{sec:evaluate_curve}, we used patch masking
to simulate feature absence. 
Instead of patch masking, we consider
common corruption~\cite{hendrycks2018benchmarking}:
fog and Gaussian noise at level 5 (for the other corruptions such
as brightness and motion blur, see Appendix~\ref{app:common}).
We apply these corruptions to image patches in the order 
selected for patch deletion in Sec.~\ref{sec:evaluate_curve}.

\begin{figure}[t]
    \begin{center}
    \includegraphics[clip, keepaspectratio, width=\linewidth]{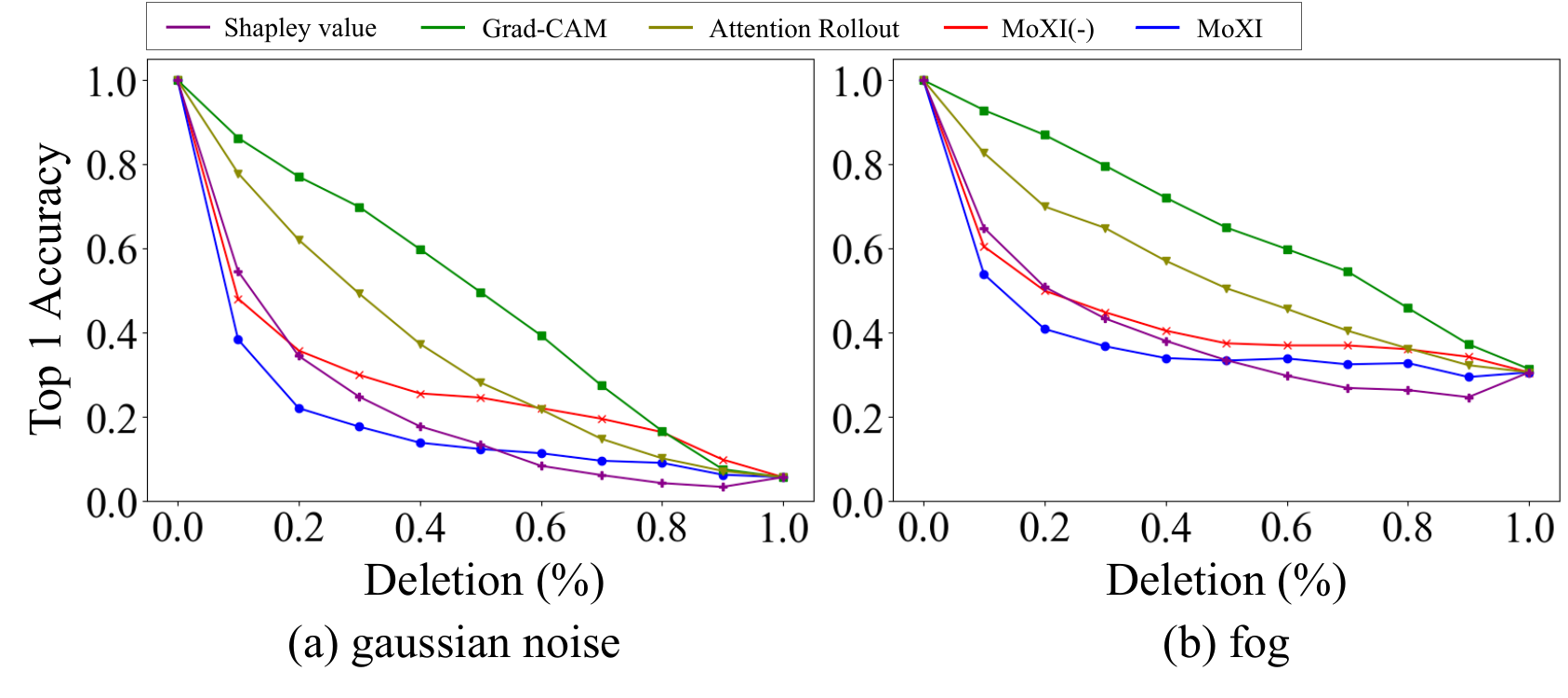}
    \caption{Deletion curves by image corruptions instead of masking: (a) Gaussian noise and (b) fog. 
    The curves illustrate the change in accuracy along with the increase in the number of corrupted image patches. The patches are corrupted from the highly contributing ones determined by each method.
    }
    \label{fig:corrupte_deletion}
    \end{center}
\end{figure}

Figure~\ref{fig:corrupte_deletion}(a) shows the effect of Gaussian noise
on the deletion curve results.
MoXI exhibits a significant decrease in accuracy compared to the others, indicating MoXI is vulnerable to Gaussian noise. This result implies that MoXI efficiently identifies important patches.
Figure~\ref{fig:corrupte_deletion}(b) shows the fog corruption results, which are similar to those observed for Gaussian noise.
Furthermore, as detailed in Appendix~\ref{app:common}, 
MoXI similarly affects accuracy with the other common corruptions.
Additionally, we evaluate the effect of adversarial perturbations. 
Interestingly, adversarial perturbations yield distinct results
due to their deceptive effect 
on the model's internal features
(see Appendix~\ref{app:adversarial}).

\subsection{Consistant explainability}\label{sec:stability}
We examine the consistent explainability of visualization methods, regardless of the internal feature representation, which is a key aspect of explainable artificial intelligence.
Specifically, we examine whether the models, trained with varying
numbers of classification classes, consistently select important image patches.
We evaluate the consistency using insertion and deletion curves
for the models trained with datasets containing 10, 20, 100, and 1000 classes.
For training the 10-class model, we select images from ImageNet that share labels with CIFAR10.
For the models with 20, 100, and 1000 classes, we extend the 10-class dataset by adding images with randomly selected classes from ImageNet. 
We draw the insertion and deletion curves using the 10-class test images that are correctly classified.

\begin{figure}[t]
    \begin{center}
    \includegraphics[clip, keepaspectratio, width=\linewidth]{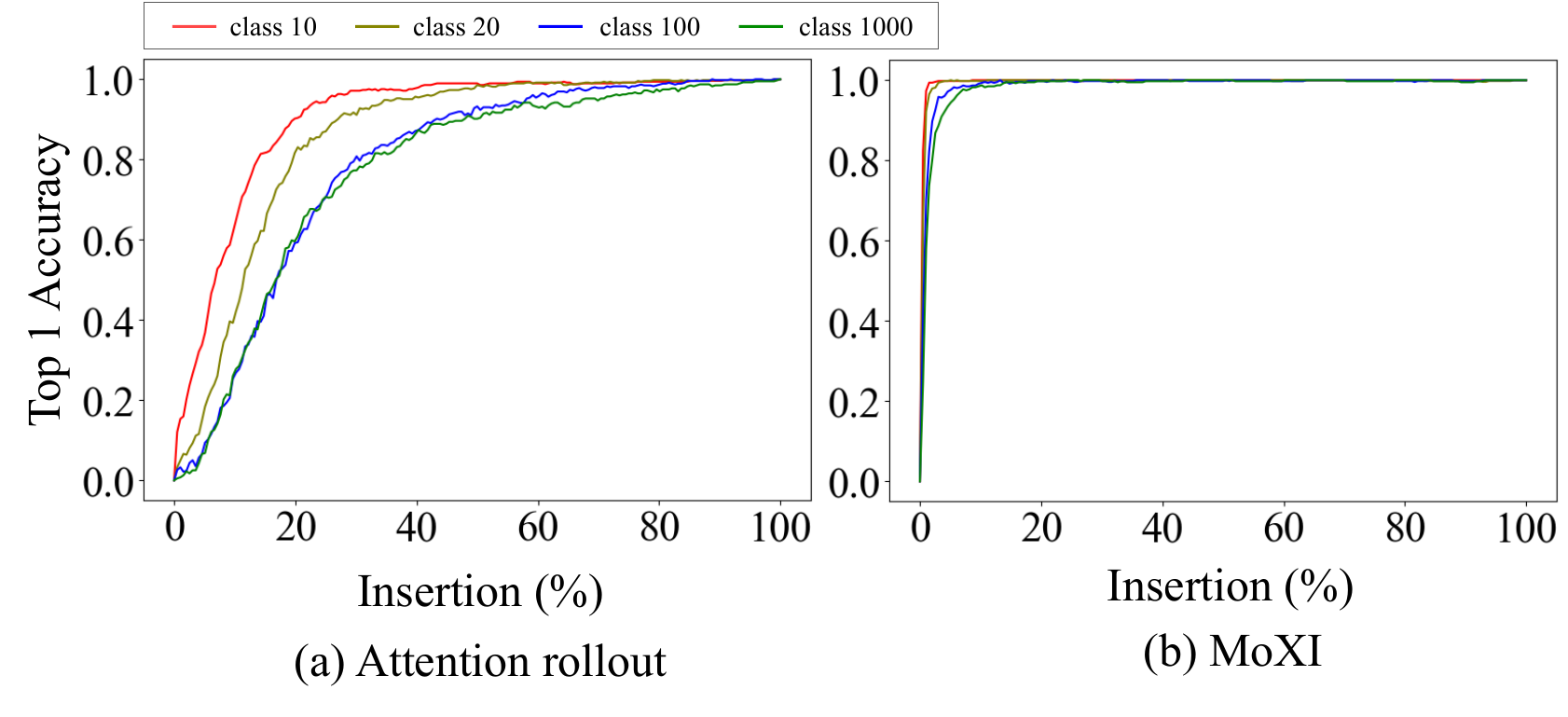}
    \caption{Insertion curves. (a) Attention rollout, (b) MoXI.
    The curves illustrate the change in accuracy along with the increase in the number of unmasked image patches.
    Each curve represents the results from the pretrained models with $10$, $20$, $100$, and $1000$ classes, respectively.
    As the number of classes the model learns increases, the accuracy of Attention rollout significantly decreases, whereas MoXI experiences only a minor decrease in accuracy.
    }
    \label{fig:class_ins_exmaple}
    \end{center}
\end{figure}

Figures~\ref{fig:class_ins_exmaple}(a) and~\ref{fig:class_ins_exmaple}(b) show the insertion curve results for Attention rollout and MoXI, respectively. 
Attention rollout decreases accuracy as the number of classes increases. 
In contrast, MoXI does not decrease in accuracy. 
Therefore, MoXI consistently selects important image patches for accurate classification.
In addition, the results from other methods and deletion experiments are shown in the 
Appendix~\ref{app:stability}. 
We confirmed that MoXI provides consistent explainability in the deletion curve experiments.

\section{Conclusion}\label{conclusion}

This study addressed the problem of identifying a group of pixels that largely and collectively impact confidence scores in image classification models. We justify simple greedy algorithms from a game-theoretic view using Shapley values and interactions. This analysis naturally suggests the use of self-context and full-context variants of Shapley values and interactions. Their computation only requires a quadratic number of forward passes, whereas prior studies compute Shapley values and/or interactions with an exponential number of forward passes or heavy sampling-based approximation.
The experimental results show that our method is more accurate in identifying the important image patches for models than popular methods. 

\section*{Acknowledgments}
This work was supported by JSPS KAKENHI Grant Number JP22H03658 and JP22K17962.

 \newcommand{\noop}[1]{}

\clearpage
\setcounter{section}{0}
\maketitlesupplementary
\renewcommand{\thesection}{\Alph{section}}

\section{Visual \textbf{\textsc{Set-Sum}} Task}\label{app:visual_setsum}
We here describe the details of the experiment for the Visual \textsc{Set-Sum} task.
The dataset consists of composite images, each of which consists of four MNIST images. The composite images are labeled by the sum of all types of digits in that image as the label (see examples in Fig.~\ref{fig:mnist_inclusion}(a)).
The size of a composite image is 56x56, and the patch size is 28x28.
As we sample the digits (i.e., MNIST images) uniformly, a composite image has duplicate numbers with a probability of roughly $47\%$. 
In the test set, each composite image was designed to have its largest digit in two patches, which is the most advantageous case of using interactions. 
We trained a ResNet-18~\cite{he2016deep} model and evaluated it on a test set of size 10,000.
The loss function used for the training is $\mathcal{L}_{\text{MNIST}} = \mathcal{L}_{\text{CE}} +  \mathcal{L}_{\text{MSE}}$, where the first loss denotes the cross-entropy loss and the second loss denotes the mean-squared error between the model prediction and the true class. 
The second loss adds a regression flavor and takes a lower value when the model prediction~(i.e., predicted set-sum) is closer to the label~(i.e., the set sum).
We filled the zero value for masking patches for computing Shapley values and interactions and for the accuracy evaluation.

\section{Results of the Insertion/Deletion curve with additional models}\label{app:deit}

\begin{figure}[t]
    \begin{center}
    \includegraphics[clip, keepaspectratio, width=\linewidth]{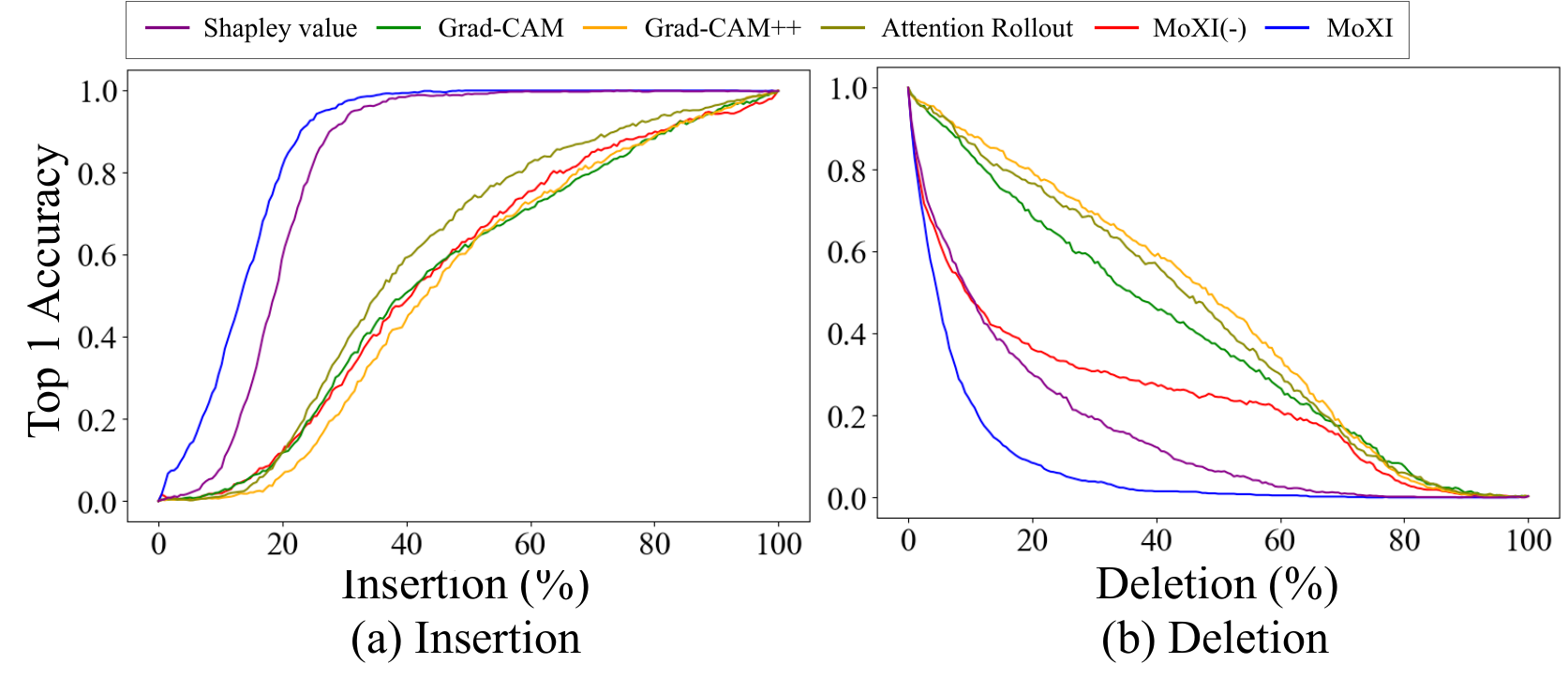}
    \caption{
    Results for DeiT-T:(a) Insertion curves. (b) Deletion curves. The curves illustrate the accuracy growth when inserting (deleting) image patches according to the contributions computed by each method. 
    }
    \label{fig:deit_imagenet_insertion_deletion}
    \end{center}
\end{figure}

\begin{figure}[t]
    \begin{center}
    \includegraphics[clip, keepaspectratio, width=\linewidth]{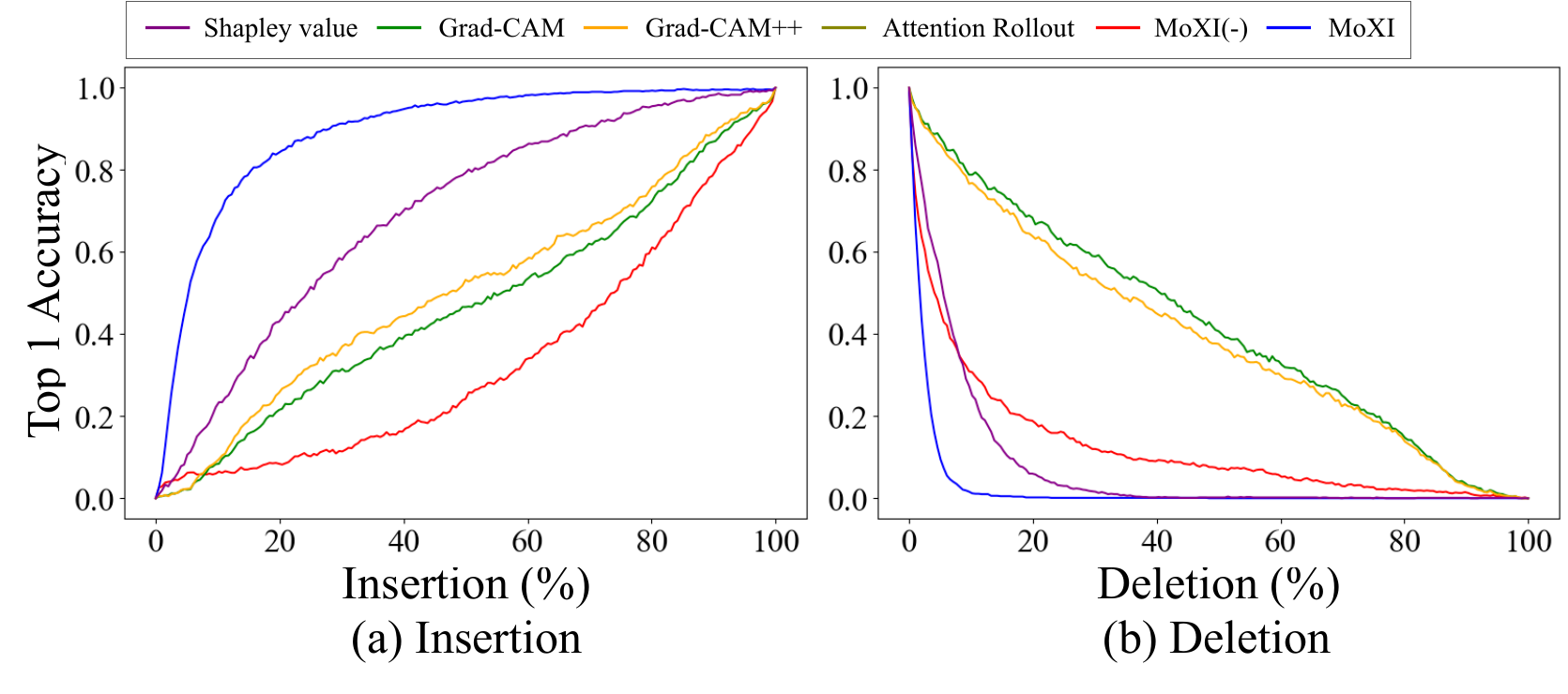}
    \caption{
    Results for ResNet-18: (a) Insertion curves. (b) Deletion curves. The curves illustrate the accuracy growth when inserting (deleting) image patches according to the contributions computed by each method. 
    }
    \label{fig:resnet_imagenet_insertion_deletion}
    \end{center}
\end{figure}

In Sec.~\ref{sec:evaluate_curve}, we evaluated the proposed and baseline methods using ViT-T~\cite{dosovitskiy2021an}. 
The insertion and deletion curves show that the proposed method provides the most efficient visual explanation.
To demonstrate this generalization across different models and architectures, we provide results using both the DeiT-T~\cite{touvron2021training}, a ViT architecture, and ResNet-18~\cite{he2016deep}, a widely used CNN model.
For details of the experiment in Deit-T, refer to Sec.~\ref{sec:evaluate_curve}. The insertion curve in Fig.~\ref{fig:deit_imagenet_insertion_deletion}(a) again shows that MoXI exhibits a sharper increase compared to the other methods.
The deletion curve in Fig.~\ref{fig:deit_imagenet_insertion_deletion}(b) also demonstrates that MoXI exhibits a sharper decrease compared to the other methods.
Similarly, Fig.~\ref{fig:resnet_imagenet_insertion_deletion} exhibits that the results for ResNet-18 are similar to these findings. 
These results indicate that our method can efficiently and accurately identify the critical patches in the model's decision-making process.

\section{Comparison of algorithm complexity and runtime}\label{app:runtime}
In this section, we compare the algorithm complexity and runtime for each method.

First, we provide an explanation of the complexity of the algorithm, focusing on the number of forward passes required per image.
Let $|N|$ be the number of patches in an image~(typically, $|N|=14^2$). Grad-CAM needs $1$ forward pass (and $1$ backward pass), Attention rollout needs $1$ forward pass, and Shapley value needs $\mathcal{O}(|N|\cdot 2^{|N|})$ forward passes. MoXI needs $\mathcal{O}(|N|^2)$ forward passes in the worst case.
The number of passes for Shapley value and MoXI is given in Sec.~\ref{sec:method}, which we will elaborate it again. 
As defined in Eq.~(1), computing the Shapley value for the $i$-th pixel requires $\mathcal{O}(2^{|N|})$ passes due to the $2^{|N|-1}$ possible choices of $S$. leading to $\mathcal{O}(|N|\cdot 2^{|N|})$ passes for an entire image. 
On the other hand, MoXI needs $O(|N|^2)$ passes. 
For example, at the $k$-th step of the greedy insertion, it recruits a new patch from the remaining $|N|-k+1$ patches to maximize the confidence score (i.e., $|N|-k+1$ passes). 
For $|N|$ steps, it needs $\mathcal{O}(|N|^2)$ passes in total. 
Note that this is the worst-case scenario; the algorithm stops when the classification becomes correct, and Fig.~\ref{fig:imagenet_inclusion_deletion}(a) indicates that more than 90\% of evaluation images require less than $0.04N$ steps. In the runtime experiment, the median of the steps was 6 (with std 7.6) and 10 (with std 11.3) for insertion and deletion, respectively.  A similar discussion holds for the deletion case.

Furthermore, our method can leverage parallel processing with mini-batches, leading to a linear number of forward passes at the cost of additional memory usage. Specifically, the $k$-th step of MoXI can be done by a single forward pass of $|N|-k+1$ patterns of the insertion from remaining patches. Our implementation is based on this parallelization.

Next, we compare the runtime required for measuring the importance in each method. 
The comparison is based on the average runtime across 100 images, following the experimental setup described in Sec.~\ref{sec:experiments}.
For Grad-CAM, Attention rollout, and Shapley value, the runtime represents the duration required to compute the importance of the entire image. In contrast, in the case of MoXI, we separately measure the runtime for pixel insertion until successful classification and for pixel deletion until classification failure.
Our experiments were conducted using a machine equipped with a 12-core processor, 64GB RAM, and an NVIDIA RTX 3090.

The runtime for each method is shown in Table~\ref{table:execution_time}. 
This indicates that the runtime for MoXI is approximately $30$ times faster than that for Shapley value.
Recalling the results from Fig.~\ref{fig:imagenet_inclusion_deletion}, MoXI achieves higher accuracy in capturing an important group of patches than Shapley value method does. 
Therefore, MoXI surpasses Shapley value in both accuracy and runtime.
While not as fast as Grad-CAM and Attention rollout, we consider that MoXI meets most use cases of visualization, and the quality is better, as our extensive experiments show.

\begin{table}[t]
\centering
\caption{Average runtime 100 ImageNet images [sec] in ViT-T. }
\label{table:execution_time}
    \begin{tabular}{|c|c|c|c|}
        \hline
        Grad-CAM & Attention R. & Shapley V. & \textbf{MoXI (Ins/Del)} \\ \hline 
        0.15 & 0.02 & 17.9 & 0.60/1.34 \\ \hline
    \end{tabular}
\end{table}

\section{Analysis of effective layers to remove patches}\label{app:mask_layer}

\begin{figure}[t]
    \begin{center}
    \includegraphics[clip, keepaspectratio, width=\linewidth]{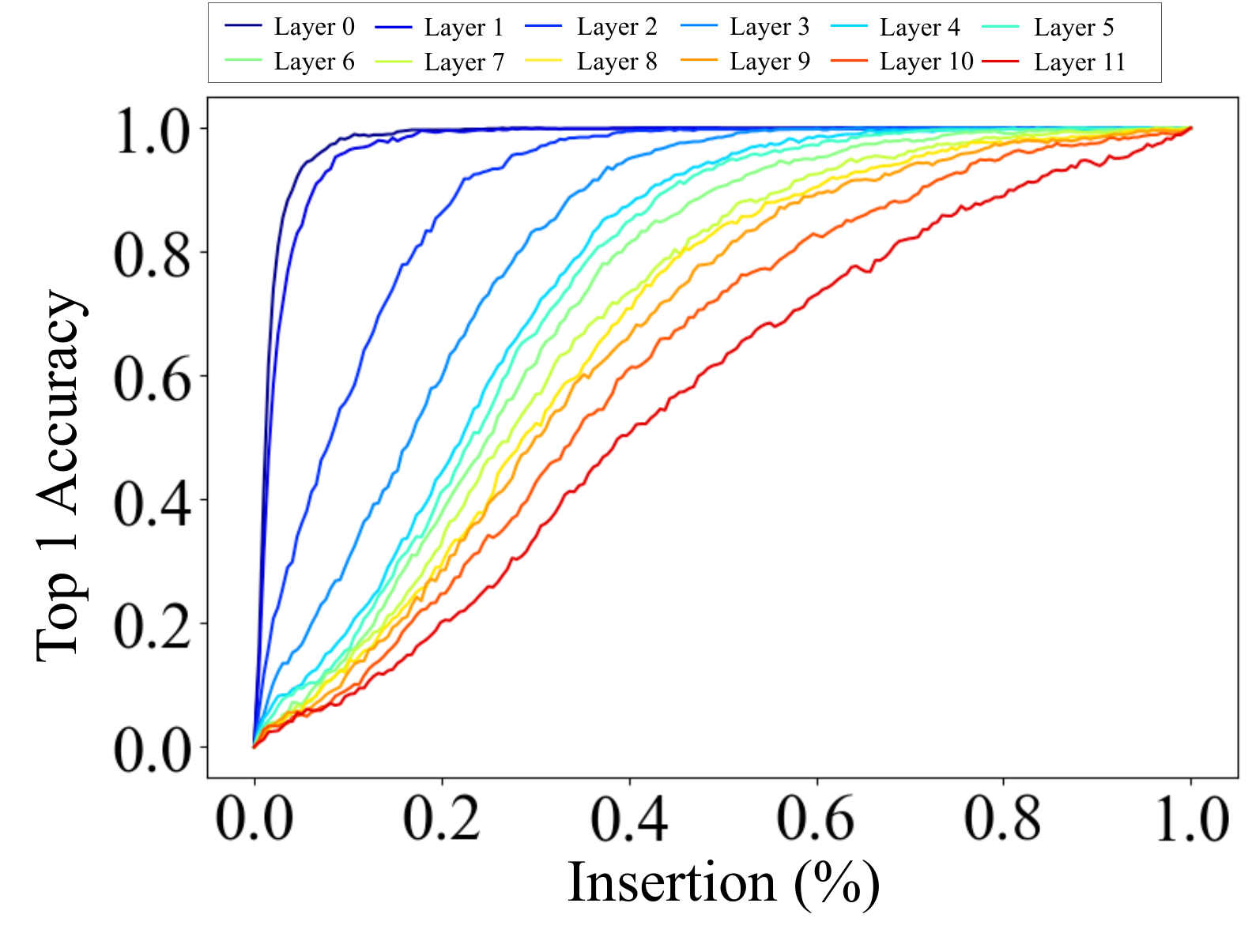}
    \caption{Insertion curves. The curves illustrate the accuracy growth when inserting image patches according to the contributions computed by each method. The horizontal axis presents the insertion rate.
    The masking method used in the computation of Shapley values and interactions employs attention masking. 
    For the insertion curve experiments, masks used for input to the model for accuracy measurement employ patch deletion.
    }
    \label{fig:layer_insertion}
    \end{center}
\end{figure}

In Sec.~\ref{sec:evaluate_curve}, we consider the absence of players (i.e., pixels/patches) for calculating Shapley values and interactions in the input space. 
Specifically, the patches are removed after the input embedding layer. 
Here, we examine the case where several self-attention layers are instead masked.
To this end, we utilize a variant of the attention-masking approach used in~\cite{covert2023learning}.
Specifically, let the $k$-th layer be our target layer. Then, a large negative value is added to the product of the query and key matrices from $k$-th to the last self-attention layers.
Figure~\ref{fig:layer_insertion} displays the insertion curve results when MoXI is applied to various target layers. The experimental setup is the same as in Sec.~\ref{sec:evaluate_curve}.
The result demonstrates that MoXI prefers the earlier layers and better pinpoints the important features of images.

\section{Additional results of visualization}\label{app:visualize_heatmp}
\begin{figure*}[t]
    \begin{center}
    \includegraphics[clip, keepaspectratio, width=\linewidth]{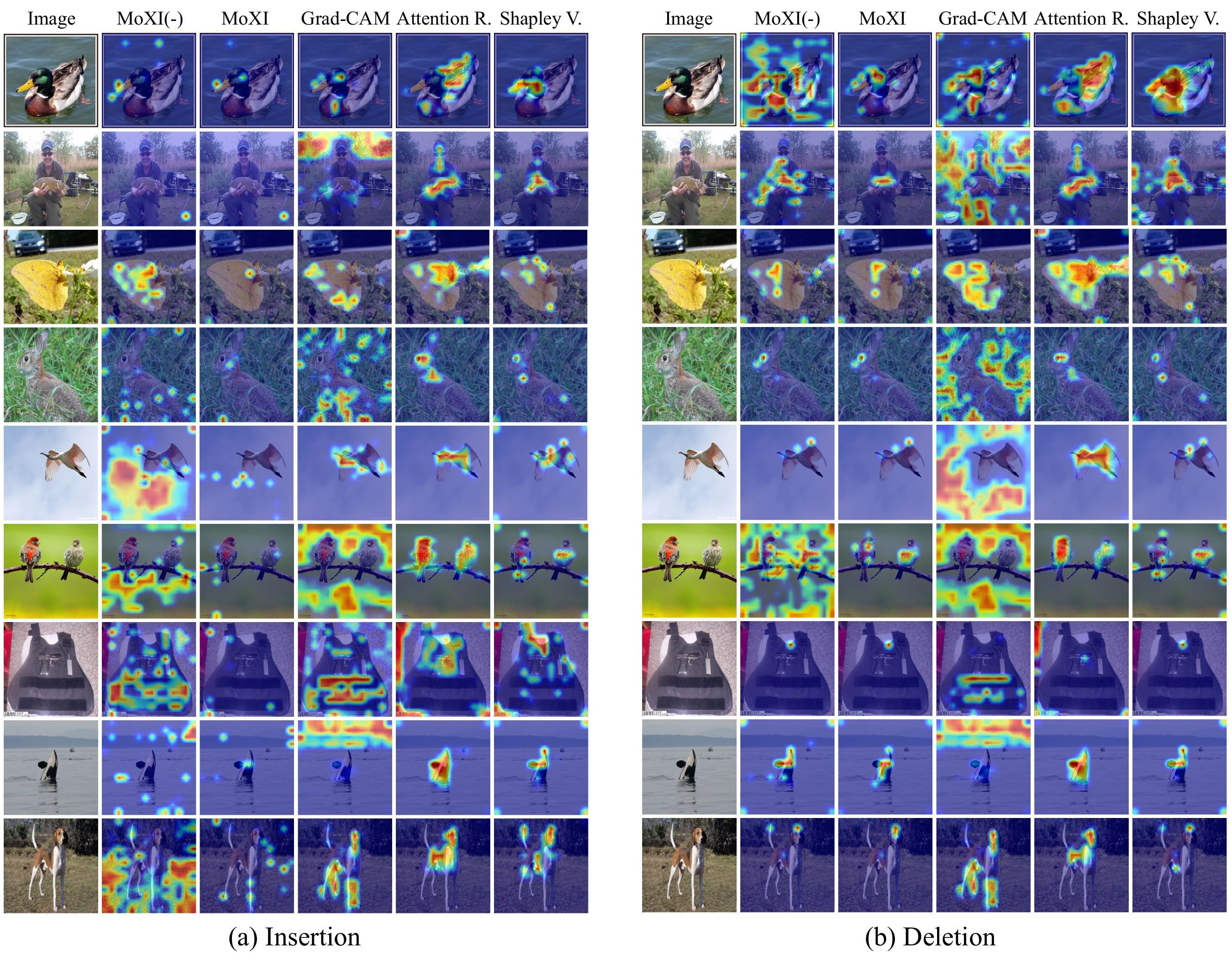}
    \caption{Visualization of important image patches by each method. 
    The highlighted image patches are selected based on their contributions calculated by each method.
    (a) Highlighting the patches incrementally added to an entire image until classification success.
    (b) Highlighting the patches sequentially removed from a full image until classification failure.
    }
    \label{fig:heatmap_large1}
    \end{center}
\end{figure*}

\begin{figure*}[t]
    \begin{center}
    \includegraphics[clip, keepaspectratio, width=\linewidth]{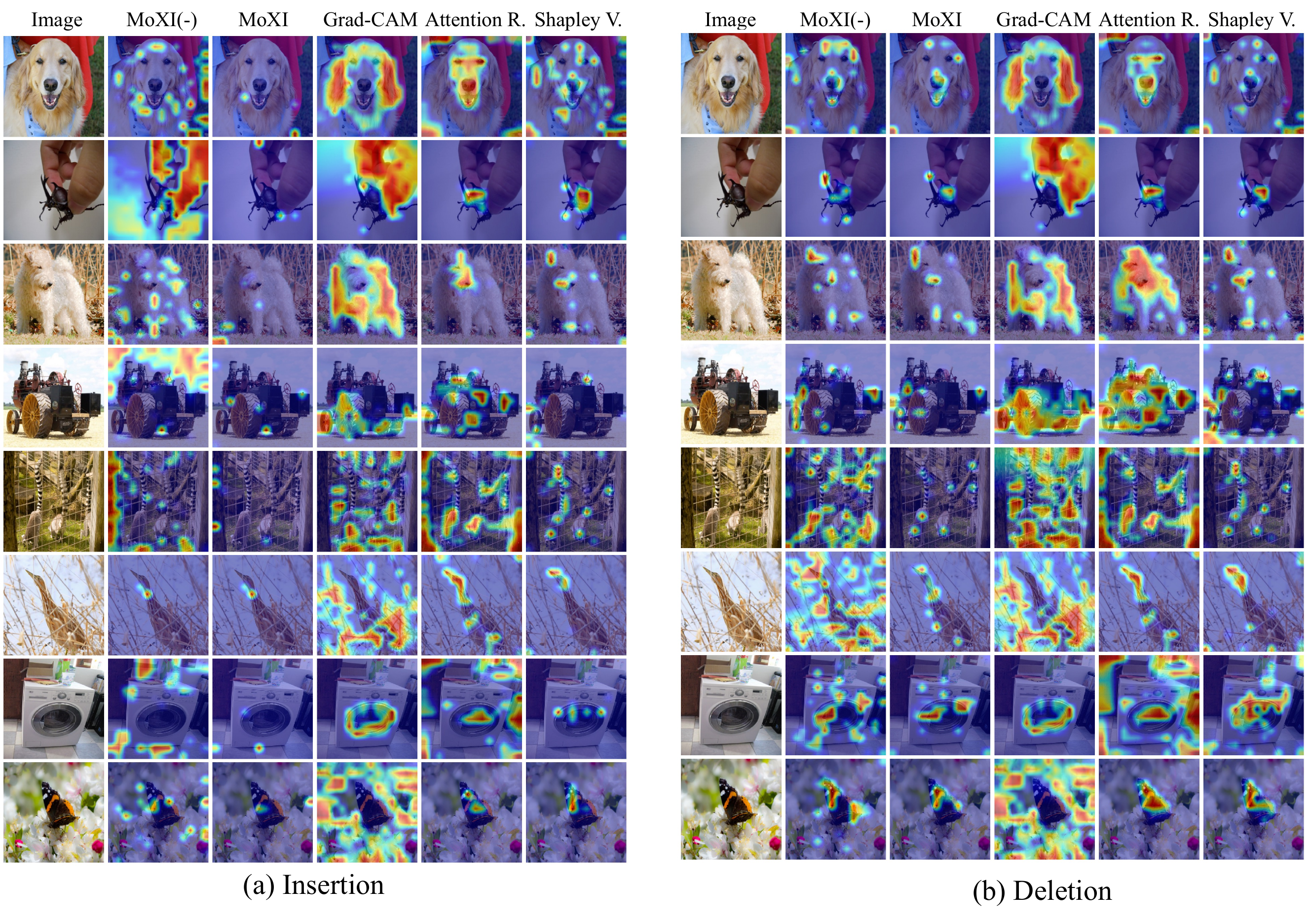}
    \caption{Visualization of important image patches by each method. 
    The highlighted image patches are selected based on their contributions calculated by each method.
    (a) Highlighting the patches incrementally added to an entire image until classification success.
    (b) Highlighting the patches sequentially removed from a full image until classification failure.
    }
    \label{fig:heatmap_large2}
    \end{center}
\end{figure*}

We provide additional visualization results in Fig.~\ref{fig:heatmap_large1} and~\ref{fig:heatmap_large2}.
As in Sec.~\ref{subsec:visualization_method}, the results demonstrated that the patches highlighted by MoXI are smaller than those highlighted by other methods.

We observed that MoXI behaves slightly unstable at the insertion case. Recall that in this case, MoXI appends important patches to an empty set accordingly and terminates when the model gives the correct classification. 
Empirically, the termination can happen at a very early stage, where the confidence score of the correct class is the largest but still very low. If we continue to patch, the model prediction can fluctuate among several classes. Note this does not cause a big problem in most cases; all the insertion curves in this paper consistently show a monotonic increase of classification accuracy with the increase in insertion rate. 
If needed, one can introduce a minimum confidence score $\tau \in [0, 1]$ and terminate the insertion when the confidence score exceeds this threshold with the correct classification. We include this hyperparameter in our official implementation of MoXI.

\section{Class-descriminative localization}\label{app:visualize_specific}

The proposed method was originally designed to identify important pixels to explain the model prediction. Here, we generalize MoXI (for pixel deletion) to visualize such pixels for a given target class, which is used in Fig.~\ref{fig:dog_and_cat}.
To this end, we consider reward function switching as follows.
Let $x, y_{\mathrm{t}}, y_{f(x)}$ be the input image, the target label, and the predicted label, respectively.
If $y_{\mathrm{t}} = y_{f(x)}$, we simply use a reward function $f(x) = \log \frac{P(y_{\mathrm{t}} \,|\, x)}{1-P(y_{\mathrm{t}} \,|\, x)}$.
Otherwise, we use $f(x) = \log \frac{P(y_{f(x)} \,|\, x)}{1-Py_{f(x)} \,|\, x)} - \frac{P(y_{\mathrm{t}} \,|\, x)}{1-P(y_{\mathrm{t}} \,|\, x)}$, which helps us identify patches with positive effect on the confidence score on class $y_{f(x)}$ and negative effect on class $y_{\mathrm{t}}$. The image patches removed in the former case are collected as important patches for class $y_{\mathrm{t}}$.

\section{Patch perturbations}\label{app:various_noise}
In Sec.~\ref{subsec:realistic}, we evaluated the effectiveness of each method by measuring the classification accuracy when Gaussian and fog noise were applied to important image patches identified. The deletion curves here are not plotted by removing patches but instead perturbed.
We present experimental results on common corruptions and adversarial perturbations.

\subsection{Common corruptions}\label{app:common}
\begin{figure*}[t]
    \begin{center}
    \includegraphics[clip, keepaspectratio, width=\linewidth]{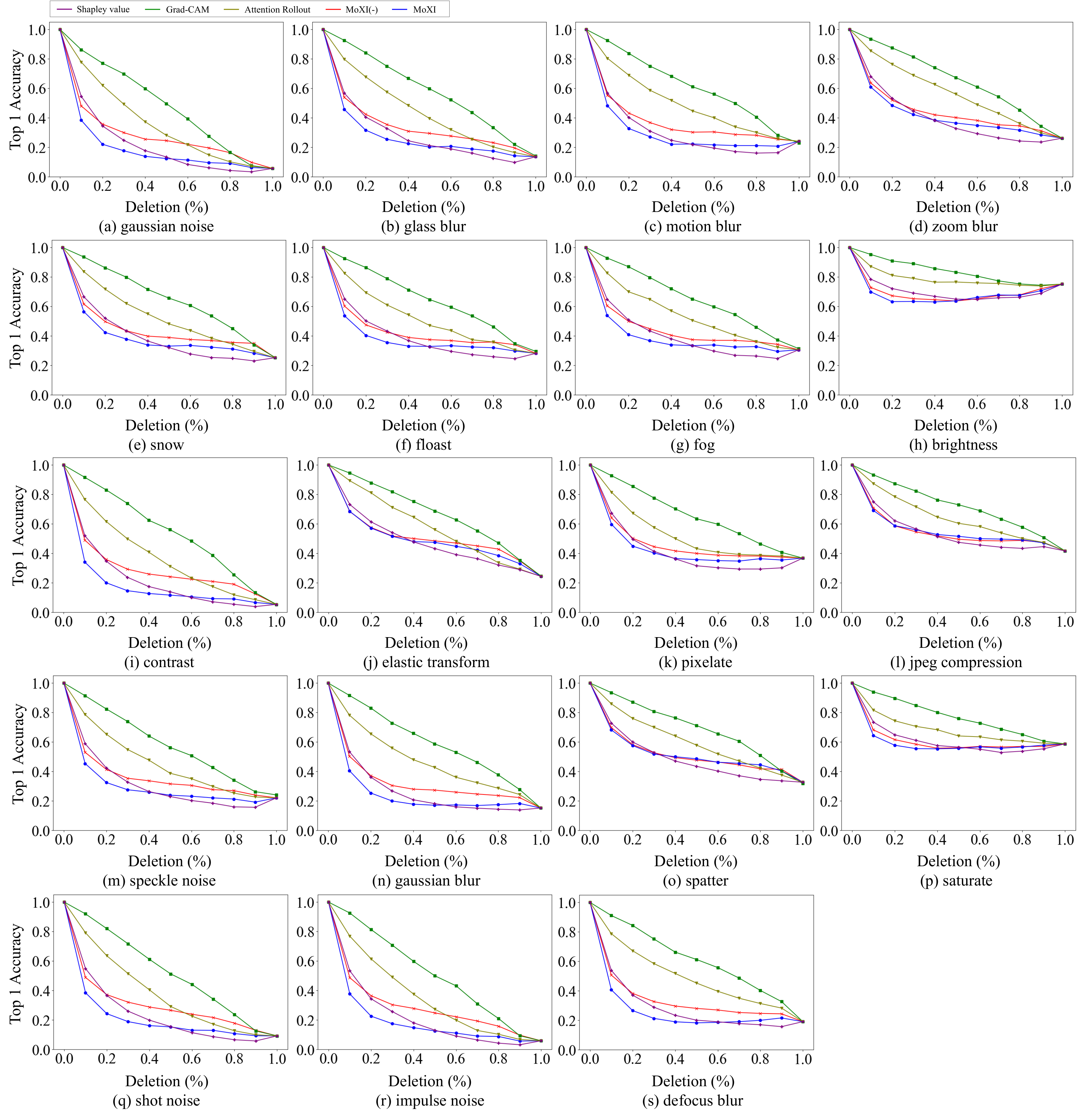}
    \caption{
    Deletion curves by image corruptions instead of masking with ViT-T.
    The curves illustrate the change in accuracy along with the increase in the number of corrupted image patches. The patches are corrupted from the highly contributing ones determined by each method.
    }
    \label{fig:corruption_all}
    \end{center}
\end{figure*}

\begin{figure*}[t]
    \begin{center}
    \includegraphics[clip, keepaspectratio, width=\linewidth]{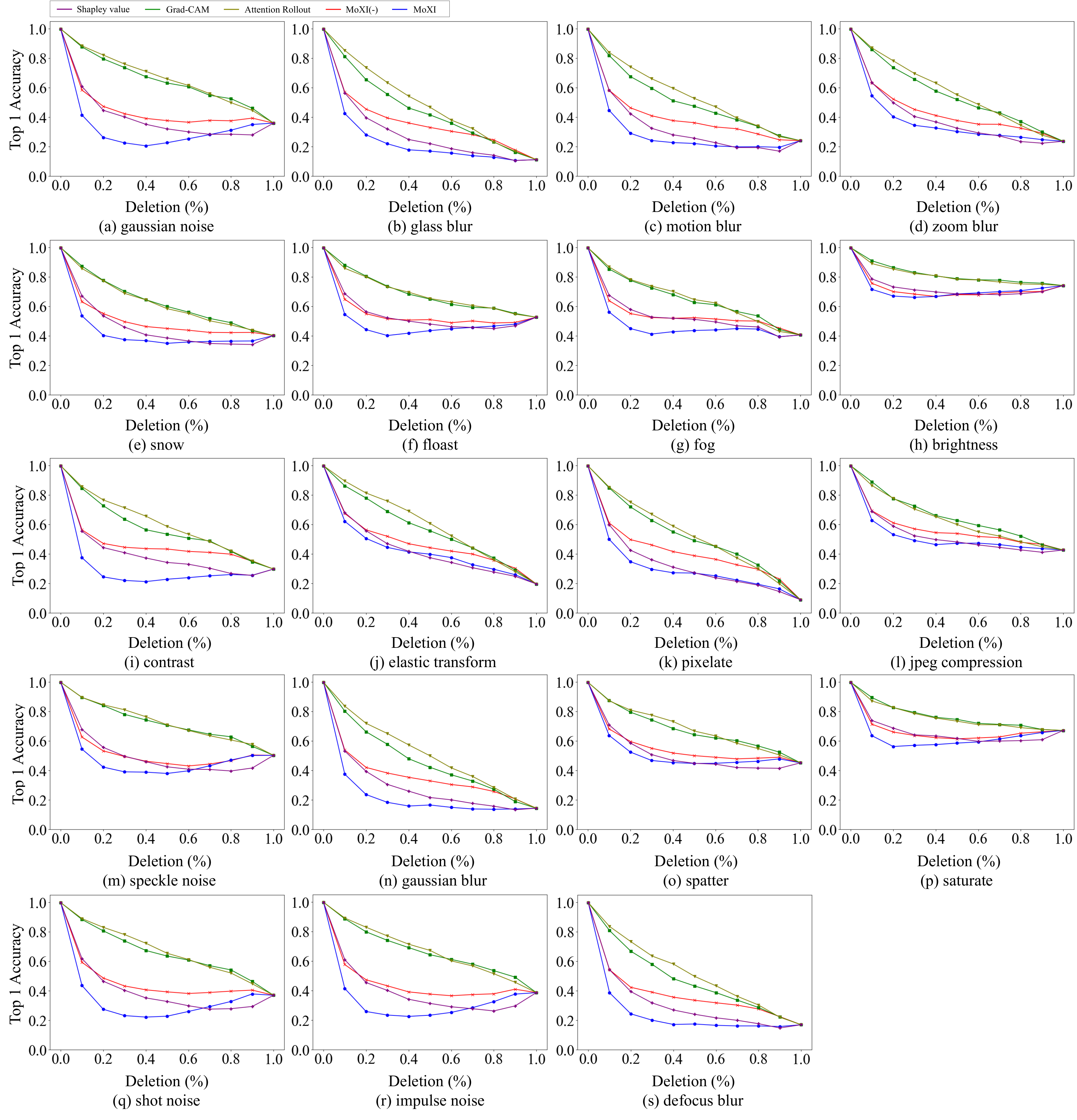}
    \caption{
    Deletion curves by image corruptions instead of masking with DeiT-T.
    The curves illustrate the change in accuracy along with the increase in the number of corrupted image patches. The patches are corrupted from the highly contributing ones determined by each method.
    }
    \label{fig:deit_corruption_all}
    \end{center}
\end{figure*}

We implemented 19 types of common corruptions using the \texttt{imagecorruptions} module with severity 5.\footnote{~\url{https://github.com/hendrycks/robustness}.}
Figures~\ref{fig:corruption_all} and~\ref{fig:deit_corruption_all} showcase the deletion curves with different corruptions for ViT-T and DeiT-T, respectively.
The results demonstrate that our method gives a sharper decrease at the early stage of deletion curves than others, as in Sec.~\ref{subsec:realistic}.

\subsection{Adversarial perturbations}\label{app:adversarial}
\begin{figure*}[t]
    \begin{minipage}{0.48\linewidth}
        \centering
        \includegraphics[clip, keepaspectratio, width=\linewidth]{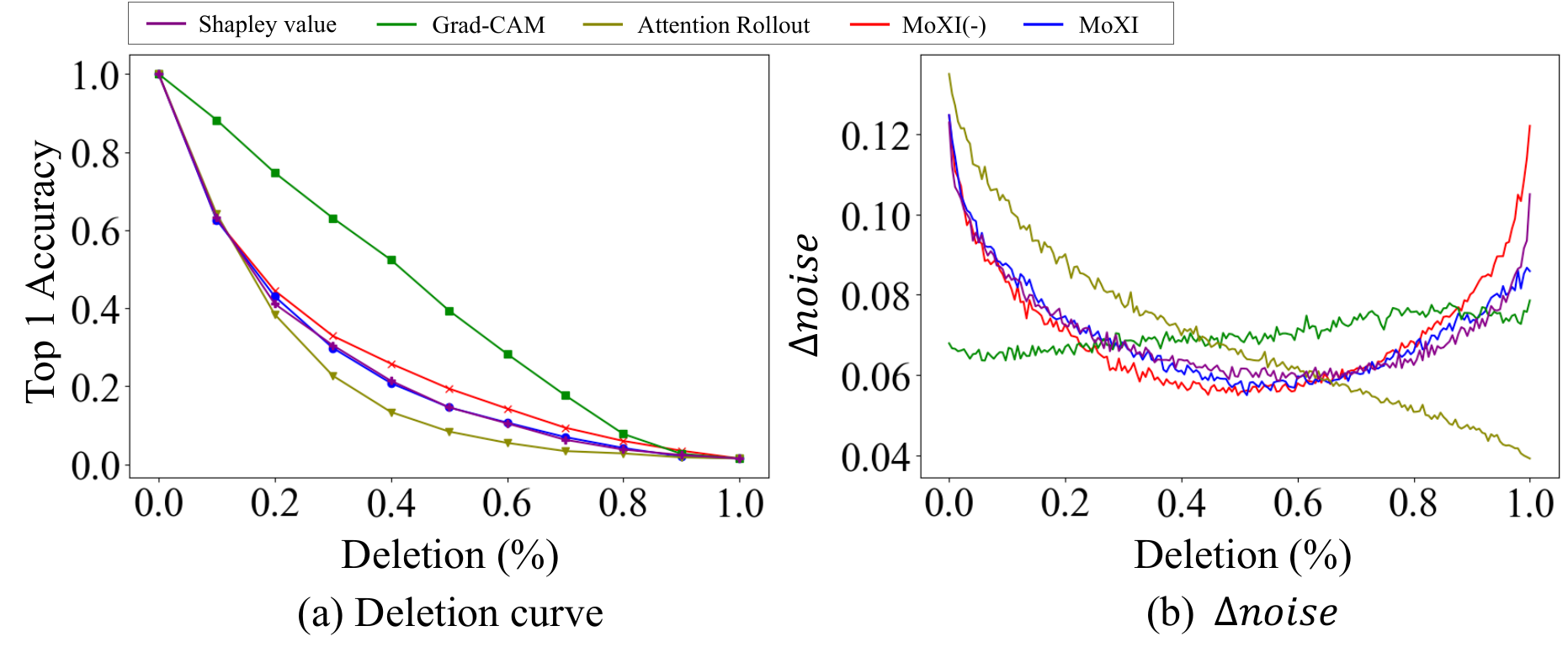}
        \caption{
        (a) 
        Deletion curves by adversarial perturbations instead of masking with ViT-T. The curves illustrate the change in accuracy along with the increase in the number of perturbed image patches. The patches are perturbed from the highly contributing ones determined by each method.
        (b) The amount of adversarial perturbations.
        }
        \label{fig:adversarial_deletion}
    \end{minipage}%
    \hfill %
    \begin{minipage}{0.48\linewidth}
        \centering
        \includegraphics[clip, keepaspectratio, width=\linewidth]{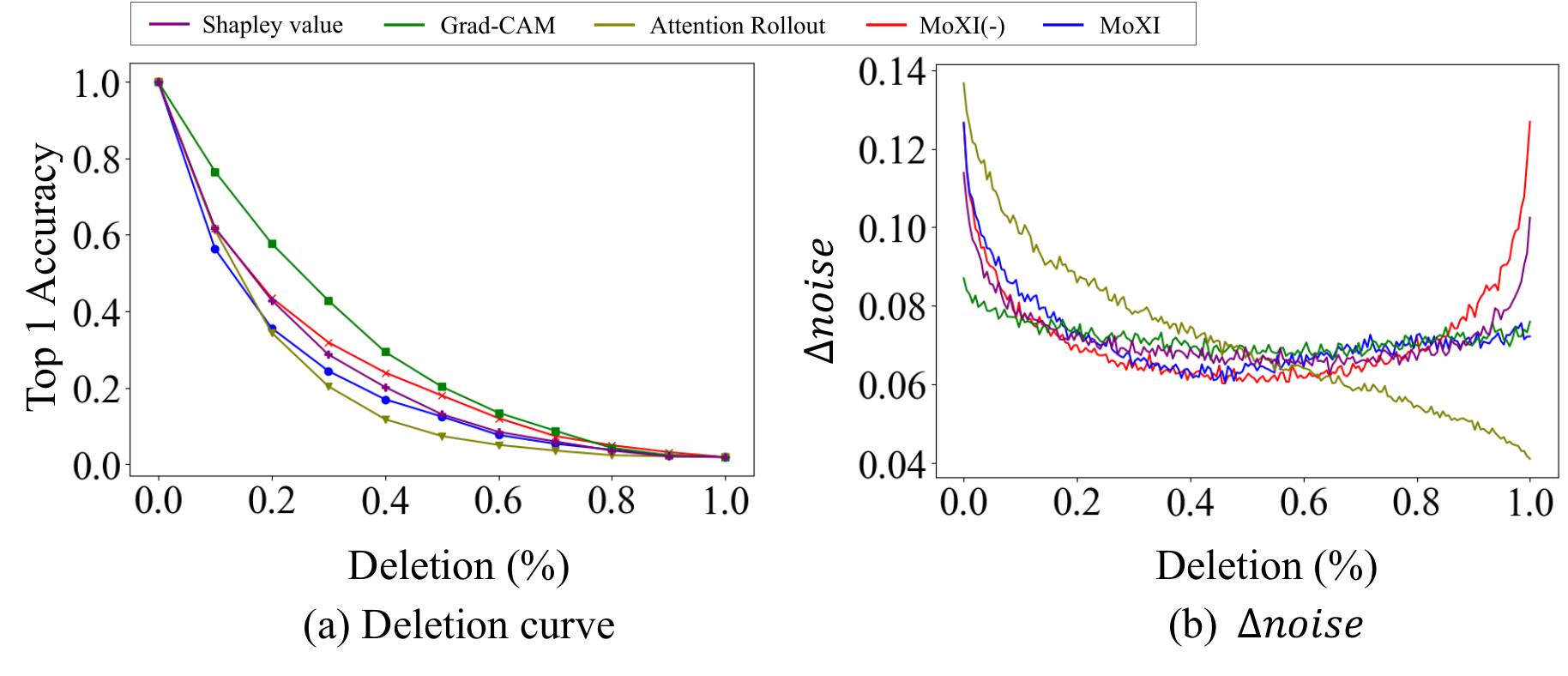}
        \caption{
        (a) Deletion curves by adversarial perturbations instead of masking with DeiT-T. The curves illustrate the change in accuracy along with the increase in the number of perturbed image patches. The patches are perturbed from the highly contributing ones determined by each method.
        (b) The amount of adversarial perturbations.
        }
        \label{fig:deit_adversarial}
    \end{minipage}
\end{figure*}

Besides common corruptions, we also investigated the case with adversarial perturbations~\cite{Goodfellow2015proceedings, alexey2017adversarial, madry2018towards}, which are small but malicious perturbations that can largely change the model's output. 
We conducted the same experiment given in Sec.~\ref{app:common} but with adversarial perturbations instead of common corruptions. 
To obtain adversarial perturbations, we adopted $L_{2}$-untargeted PGD with $\epsilon=1.0$ and stepsize $\alpha=0.2$.
Figure~\ref{fig:adversarial_deletion}(a) and~\ref{fig:deit_adversarial}(a) present the deletion curves for ViT-T and Deit-T, respectively.
The results show that the attention rollout method gives a slightly sharper decrease than MoXI. 
This differs from the results for common corruptions. 
We suspect that adversarial perturbations mostly lie in the patches that are suggested as important by attention rollout. 
To confirm this, we measured the magnitude of adversarial perturbations on each image patch. 
Specifically, the magnitude is measured by the L2 norm. 
Figure~\ref{fig:adversarial_deletion}(b) shows the magnitude of the perturbations of each patch. 
The patches are ordered as in the deletion curves in Fig.~\ref{fig:adversarial_deletion}(a).
The results indicate that the importance of image patches identified by attention rollout is well aligned with the amount of perturbations on them. 
On the other hand, image patches identified by MoXI contain a larger amount of perturbations at the early and late stages than those at the middle stage. 
This may be because the attention rollout reflects the internal computation process of the features directly when measuring the contributions of image patches, while adversarial perturbations are designed to hack this process. On the other hand, MoXI treats a Vision Transformer as a black-box model and is unaware of the internal process.

\section{More results in the stability of explanations.}\label{app:stability}
\begin{figure*}[t]
    \begin{center}
    \includegraphics[clip, keepaspectratio, width=\linewidth]{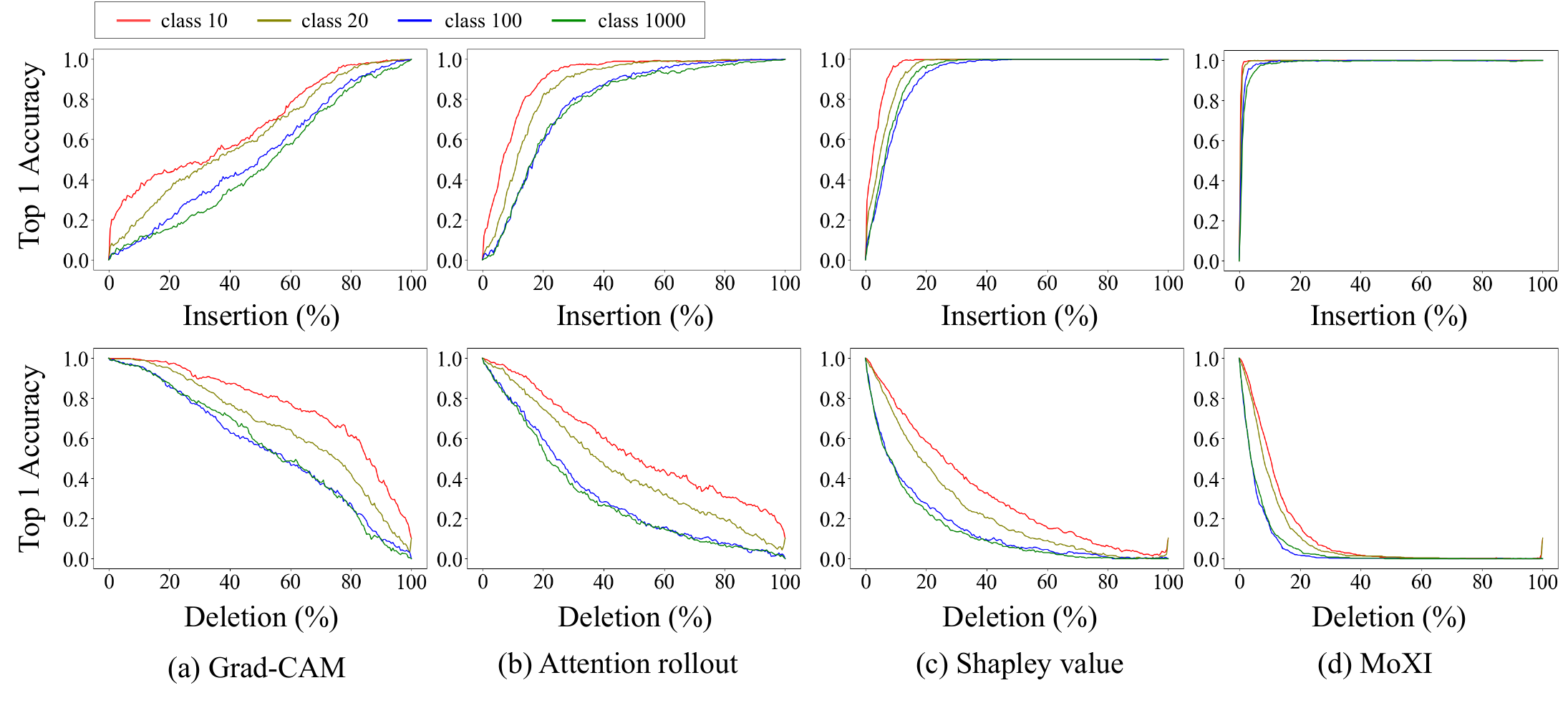}
    \caption{
    (Top) Insertion curves. (Bottom) Deletion curves. 
    The curves illustrate the change in accuracy along with the increase (decrease) in the number of unmasked (masked) image patches.
    Each curve represents the results from the pretrained models with $10$, $20$, $100$, and $1000$ classes, respectively.
    (a) Grad-CAM results, (b) Attention Rollout results, (c) Shapley Value results, (d) MoXI results.
    }
    \label{fig:class_deletion_all}
    \end{center}
\end{figure*}

In Sec~\ref{sec:stability}, we evaluate the stability of explanations of MoXI and attention rollout with respect to the number of classes.
Here, we consider both insertion and deletion metrics, utilizing Grad-CAM, attention rollout, Shapley value, and MoXI.
Figure~\ref{fig:class_deletion_all} shows insertion and deletion curves.
The result again shows that MoXI maintains relatively stable accuracy when the model is trained on more classes.
Similarly, other methods have significantly decreased classification accuracy in such scenarios.
Therefore, MoXI acquires important image patches more consistently than other methods.

\begin{thebibliography}{33}
\providecommand{\natexlab}[1]{#1}
\providecommand{\url}[1]{\texttt{#1}}
\expandafter\ifx\csname urlstyle\endcsname\relax
  \providecommand{\doi}[1]{doi: #1}\else
  \providecommand{\doi}{doi: \begingroup \urlstyle{rm}\Url}\fi

\bibitem[Abnar and Zuidema(2020)]{abnar2020quantifying}
Samira Abnar and Willem Zuidema.
\newblock Quantifying attention flow in transformers.
\newblock In \emph{Proceedings of the 58th Annual Meeting of the Association for Computational Linguistics}, pages 4190--4197, Online, 2020. Association for Computational Linguistics.

\bibitem[Ancona et~al.(2019)Ancona, Oztireli, and Gross]{ancona19a}
Marco Ancona, Cengiz Oztireli, and Markus Gross.
\newblock Explaining deep neural networks with a polynomial time algorithm for shapley value approximation.
\newblock In \emph{Proceedings of the 36th International Conference on Machine Learning}, pages 272--281, Long Beach, California, USA, 2019. PMLR.

\bibitem[Binder et~al.(2016)Binder, Montavon, Lapuschkin, M{\"{u}}ller, and Samek]{Binder2016layer}
Alexander Binder, Gr{\'{e}}goire Montavon, Sebastian Lapuschkin, Klaus-Robert M{\"{u}}ller, and Wojciech Samek.
\newblock \emph{Layer-Wise Relevance Propagation for Neural Networks with Local Renormalization Layers}, pages 63--71.
\newblock Springer International Publishing, Cham, 2016.

\bibitem[Bl\"ucher et~al.(2022)Bl\"ucher, Vielhaben, and Strodthoff]{blucher2022prediff}
Stefan Bl\"ucher, Johanna Vielhaben, and Nils Strodthoff.
\newblock {PredDiff}: {E}xplanations and interactions from conditional expectations.
\newblock \emph{Artificial Intelligence}, 312:\penalty0 103774, 2022.

\bibitem[Bluecher et~al.(2024)Bluecher, Vielhaben, and Strodthoff]{bluecher2024decoupling}
Stefan Bluecher, Johanna Vielhaben, and Nils Strodthoff.
\newblock Decoupling pixel flipping and occlusion strategy for consistent {XAI} benchmarks.
\newblock \emph{Transactions on Machine Learning Research}, 2024.

\bibitem[Castro et~al.(2009)Castro, Gómez, and Tejada]{CASTRO2009polynomial}
Javier Castro, Daniel Gómez, and Juan Tejada.
\newblock Polynomial calculation of the shapley value based on sampling.
\newblock \emph{Computers \& Operations Research}, 36\penalty0 (5):\penalty0 1726--1730, 2009.
\newblock Selected papers presented at the Tenth International Symposium on Locational Decisions (ISOLDE X).

\bibitem[Chattopadhay et~al.(2018)Chattopadhay, Sarkar, Howlader, and Balasubramanian]{chattopadhyay2018grad_cam}
Aditya Chattopadhay, Anirban Sarkar, Prantik Howlader, and Vineeth~N Balasubramanian.
\newblock Grad-{CAM}$++$: Generalized gradient-based visual explanations for deep convolutional networks.
\newblock In \emph{Proceedings of the IEEE Winter Conference on Applications of Computer Vision}. {IEEE}, 2018.

\bibitem[Chefer et~al.(2021)Chefer, Gur, and Wolf]{Chefer2021TransformerInt}
Hila Chefer, Shir Gur, and Lior Wolf.
\newblock Transformer interpretability beyond attention visualization.
\newblock In \emph{Proceedings of the IEEE Conference on Computer Vision and Pattern Recognition}, pages 782--791, 2021.

\bibitem[Cheng et~al.(2021)Cheng, Chu, Zheng, Ren, and Zhang]{cheng2021agame}
Xu Cheng, Chuntung Chu, Yi Zheng, Jie Ren, and Quanshi Zhang.
\newblock A game-theoretic taxonomy of visual concepts in {DNNs}.
\newblock \emph{arXiv preprint arXiv:2106.10938}, 2021.

\bibitem[Covert et~al.(2023)Covert, Kim, and Lee]{covert2023learning}
Ian~Connick Covert, Chanwoo Kim, and Su-In Lee.
\newblock Learning to estimate shapley values with vision transformers.
\newblock In \emph{The Eleventh International Conference on Learning Representations}, 2023.

\bibitem[Deng et~al.(2022)Deng, Ren, Zhang, and Zhang]{deng2022discovering}
Huiqi Deng, Qihan Ren, Hao Zhang, and Quanshi Zhang.
\newblock Discovering and explaining the representation bottleneck of {DNNs}.
\newblock In \emph{Proceedings of the International Conference on Learning Representations}, 2022.

\bibitem[Deng et~al.(2009)Deng, Dong, Socher, Li, Li, and Fei-Fei]{deng2009imagenet}
Jia Deng, Wei Dong, Richard Socher, Li-Jia Li, Kai Li, and Li Fei-Fei.
\newblock {ImageNet}: A large-scale hierarchical image database.
\newblock In \emph{Proceedings of the IEEE Conference on Computer Vision and Pattern Recognition}, pages 248--255, 2009.

\bibitem[Dosovitskiy et~al.(2021)Dosovitskiy, Beyer, Kolesnikov, Weissenborn, Zhai, Unterthiner, Dehghani, Minderer, Heigold, Gelly, Uszkoreit, and Houlsby]{dosovitskiy2021an}
Alexey Dosovitskiy, Lucas Beyer, Alexander Kolesnikov, Dirk Weissenborn, Xiaohua Zhai, Thomas Unterthiner, Mostafa Dehghani, Matthias Minderer, Georg Heigold, Sylvain Gelly, Jakob Uszkoreit, and Neil Houlsby.
\newblock An image is worth 16x16 words: Transformers for image recognition at scale.
\newblock In \emph{Proceedings of the International Conference on Learning Representations}, 2021.

\bibitem[Goodfellow et~al.(2015)Goodfellow, Shlens, and Szegedy]{Goodfellow2015proceedings}
Ian Goodfellow, Jonathon Shlens, and Christian Szegedy.
\newblock Explaining and harnessing adversarial examples.
\newblock In \emph{Proceedings of the International Conference on Learning Representations}, 2015.

\bibitem[Grabisch and Roubens(1999)]{grabisch1999an}
Michel Grabisch and Marc Roubens.
\newblock An axiomatic approach to the concept of interaction among players in cooperative games.
\newblock \emph{International Journal of Game Theory}, 28:\penalty0 547--565, 1999.

\bibitem[He et~al.(2016)He, Zhang, Ren, and Sun]{he2016deep}
Kaiming He, Xiangyu Zhang, Shaoqing Ren, and Jian Sun.
\newblock Deep residual learning for image recognition.
\newblock In \emph{Proceedings of the IEEE Conference on Computer Vision and Pattern Recognition}, pages 770--778, 2016.

\bibitem[Hendrycks and Dietterich(2019)]{hendrycks2018benchmarking}
Dan Hendrycks and Thomas Dietterich.
\newblock Benchmarking neural network robustness to common corruptions and perturbations.
\newblock In \emph{Proceedings of the International Conference on Learning Representations}, 2019.

\bibitem[Jethani et~al.(2022)Jethani, Sudarshan, Covert, Lee, and Ranganath]{jethani2022fastshap}
Neil Jethani, Mukund Sudarshan, Ian~Connick Covert, Su-In Lee, and Rajesh Ranganath.
\newblock Fast{SHAP}: Real-time {Shapley} value estimation.
\newblock In \emph{Proceedings of the International Conference on Learning Representations}, 2022.

\bibitem[Kurakin et~al.(2017)Kurakin, Goodfellow, and Bengio]{alexey2017adversarial}
Alexey Kurakin, Ian Goodfellow, and Samy Bengio.
\newblock Adversarial examples in the physical world.
\newblock \emph{Proceedings of the International Conference on Learning Representations Workshop}, 2017.

\bibitem[Lundberg and Lee(2017)]{Lundberg2017aunified}
Scott~M. Lundberg and Su-In Lee.
\newblock A unified approach to interpreting model predictions.
\newblock In \emph{Proceedings of the 31st International Conference on Neural Information Processing Systems}, page 4768–4777, Red Hook, NY, USA, 2017. Curran Associates Inc.

\bibitem[Madry et~al.(2018)Madry, Makelov, Schmidt, Tsipras, and Vladu]{madry2018towards}
Aleksander Madry, Aleksandar Makelov, Ludwig Schmidt, Dimitris Tsipras, and Adrian Vladu.
\newblock Towards deep learning models resistant to adversarial attacks.
\newblock In \emph{Proceedings of the International Conference on Learning Representations}, 2018.

\bibitem[Petsiuk et~al.(2018)Petsiuk, Das, and Saenko]{Petsiuk2018rise}
Vitali Petsiuk, Abir Das, and Kate Saenko.
\newblock Rise: Randomized input sampling for explanation of black-box models.
\newblock In \emph{Proceedings of the British Machine Vision Conference}, 2018.

\bibitem[Ren et~al.(2021)Ren, Zhang, Wang, Chen, Zhou, Chen, Cheng, Wang, Zhou, Shi, and Zhang]{ren2021a}
Jie Ren, Die Zhang, Yisen Wang, Lu Chen, Zhanpeng Zhou, Yiting Chen, Xu Cheng, Xin Wang, Meng Zhou, Jie Shi, and Quanshi Zhang.
\newblock Towards a unified game-theoretic view of adversarial perturbations and robustness.
\newblock In \emph{Proceedings of the Advances in Neural Information Processing Systems}, pages 3797--3810, 2021.

\bibitem[Ren et~al.(2022)Ren, Zhou, Chen, and Zhang]{ren2022towards}
Jie Ren, Zhanpeng Zhou, Qirui Chen, and Quanshi Zhang.
\newblock Towards a game-theoretic view of baseline values in the shapley value, 2022.

\bibitem[Selvaraju et~al.(2017)Selvaraju, Cogswell, Das, Vedantam, Parikh, and Batra]{Selvaraju2018Grad_CAM}
Ramprasaath~R. Selvaraju, Michael Cogswell, Abhishek Das, Ramakrishna Vedantam, Devi Parikh, and Dhruv Batra.
\newblock {Grad-CAM}: Visual explanations from deep networks via gradient-based localization.
\newblock In \emph{Proceedings of IEEE/CVF International Conference on Computer Vision}, pages 618--626, 2017.

\bibitem[Shapley(1953)]{shapley1953contibutions}
Lloyd~S. Shapley.
\newblock A value for n-person games.
\newblock In \emph{Contributions to the Theory of Games}, pages 307--317, 1953.

\bibitem[Sumiyasu et~al.(2022)Sumiyasu, Kawamoto, and Kera]{sumiyasu2022gametheoretic}
Kosuke Sumiyasu, Kazuhiko Kawamoto, and Hiroshi Kera.
\newblock Game-theoretic understanding of misclassification, 2022.

\bibitem[Touvron et~al.(2021)Touvron, Cord, Douze, Massa, Sablayrolles, and Jegou]{touvron2021training}
Hugo Touvron, Matthieu Cord, Matthijs Douze, Francisco Massa, Alexandre Sablayrolles, and Herve Jegou.
\newblock Training data-efficient image transformers \& distillation through attention.
\newblock In \emph{Proceedings of the 38th International Conference on Machine Learning}, pages 10347--10357, 2021.

\bibitem[Wang et~al.(2020)Wang, Wang, Du, Yang, Zhang, Ding, Mardziel, and Hu]{wang2020scorecam}
Haofan Wang, Zifan Wang, Mengnan Du, Fan Yang, Zijian Zhang, Sirui Ding, Piotr Mardziel, and Xia Hu.
\newblock Score-cam: Score-weighted visual explanations for convolutional neural networks, 2020.

\bibitem[Wang et~al.(2021)Wang, Ren, Lin, Zhu, Wang, and Zhang]{wang2021a}
Xin Wang, Jie Ren, Shuyun Lin, Xiangming Zhu, Yisen Wang, and Quanshi Zhang.
\newblock A unified approach to interpreting and boosting adversarial transferability.
\newblock In \emph{Proceedings of the International Conference on Learning Representations}, 2021.

\bibitem[Zhang et~al.(2021{\natexlab{a}})Zhang, Li, Ma, Li, Xie, and Zhang]{zhang2021interpreting}
Hao Zhang, Sen Li, YinChao Ma, Mingjie Li, Yichen Xie, and Quanshi Zhang.
\newblock Interpreting and boosting dropout from a game-theoretic view.
\newblock In \emph{Proceedings of the International Conference on Learning Representations}, 2021{\natexlab{a}}.

\bibitem[Zhang et~al.(2021{\natexlab{b}})Zhang, Xie, Zheng, Zhang, and Zhang]{zhang2020Interpreting}
Hao Zhang, Yichen Xie, Longjie Zheng, Die Zhang, and Quanshi Zhang.
\newblock Interpreting multivariate shapley interactions in dnns.
\newblock In \emph{The AAAI Conference on Artificial Intelligence}, 2021{\natexlab{b}}.

\bibitem[Zhou et~al.(2016)Zhou, Khosla, Lapedriza, Oliva, and Torralba]{zhou2016learning}
Bolei Zhou, Aditya Khosla, Agata Lapedriza, Aude Oliva, and Antonio Torralba.
\newblock Learning deep features for discriminative localization.
\newblock In \emph{Proceedings of the IEEE Conference on Computer Vision and Pattern Recognition}, pages 2921--2929, 2016.

\end{thebibliography}
\end{document}